%% file: main.tex
\documentclass{article}

\usepackage[final]{neurips_2022}

\usepackage[utf8]{inputenc} 
\usepackage[T1]{fontenc}    
\usepackage{url}            
\usepackage{booktabs}       
\usepackage{amsfonts}       
\usepackage{nicefrac}       
\usepackage{microtype}      
\usepackage{xcolor}         
\usepackage{graphicx}
\usepackage{amsmath,amsfonts,amssymb}
\usepackage{algorithm}
\usepackage{enumitem}
\usepackage{natbib}
\setcitestyle{square,numbers}

\usepackage{comment}
\usepackage{multirow}
\usepackage{wrapfig}
\usepackage{makecell}
\usepackage[noend]{algcompatible}
\usepackage{arydshln} 

\usepackage[pagebackref=true,breaklinks=true,letterpaper=true,colorlinks,bookmarks=false]{hyperref}

\newcommand{\nickname}{OGC}
\newcommand{\etal}{\textit{et al}.}
\newcommand{\ie}{\textit{i}.\textit{e}.}

\newcommand{\etc}{\textit{etc}.}

\usepackage{xcolor}
\usepackage{color,soul}

\title{\nickname{}: Unsupervised 3D Object Segmentation from Rigid Dynamics of Point Clouds}

\author{%
   Ziyang Song \quad Bo Yang \vspace{0.2cm} \\
   vLAR Group, The Hong Kong Polytechnic University \\
   {\tt \small ziyang.song@connect.polyu.hk \quad bo.yang@polyu.edu.hk}
}

\begin{document}
\maketitle

\begin{abstract}
    In this paper, we study the problem of 3D object segmentation from raw point clouds. Unlike all existing methods which usually require a large amount of human annotations for full supervision, we propose the first unsupervised method, called \nickname{}, to simultaneously identify multiple 3D objects in a single forward pass, without needing any type of human annotations. The key to our approach is to fully leverage the dynamic motion patterns over sequential point clouds as supervision signals to automatically discover rigid objects. Our method consists of three major components, 1) the object segmentation network to directly estimate multi-object masks from a single point cloud frame, 2) the auxiliary self-supervised scene flow estimator, and 3) our core object geometry consistency component. By carefully designing a series of loss functions, we effectively take into account the multi-object rigid consistency and the object shape invariance in both temporal and spatial scales. This allows our method to truly discover the object geometry even in the absence of annotations. We extensively evaluate our method on five datasets, demonstrating the superior performance for object part instance segmentation and general object segmentation in both indoor and the challenging outdoor scenarios. 
    Our code and data are available at \url{https://github.com/vLAR-group/OGC} 
\end{abstract}

\section{Introduction}
\input{chaps/01_intro}

\section{Related Works}
\input{chaps/02_liter}

\section{\nickname{}}
\input{chaps/03_meth}

\section{Experiments}
\input{chaps/04_exp}

\section{Conclusion}
\input{chaps/05_sum}

\clearpage
{\small
\bibliographystyle{abbrv}
\bibliography{references}
}

\clearpage
\include{chaps/06_checklist}

\clearpage
\appendix
\section{Appendix}
\input{chaps/07_app}

\end{document}

%% file: chaps/01_intro.tex
Identifying 3D objects from point clouds is vital for machines to tackle high-level tasks such as autonomous planning and manipulation in real-world scenarios. Inspired by the seminal work PointNet \cite{Qi2016}, a plethora of sophisticated models \cite{Wang2018d,Yang2019d,Lang2019} have been developed to accurately detect and segment individual objects from the sparse and irregular point clouds. Although these methods have achieved excellent performance on a wide range of public datasets, they primarily rely on a huge amount of human annotations for full supervision. However, it is extremely costly to fully annotate every objects in point clouds due to the irregularity of data format. 

Very recently, a few works start to address 3D object segmentation in the absence of human annotations. By analysing 3D scene flows from sequential point clouds, Jiang \etal{} \cite{Jiang2021} apply the conventional subspace clustering optimization technique to identify moving objects from raw point cloud sequences. With the self-supervised learning of 3D scene flow, SLIM \cite{Baur2021} is the first learning-based work to showcase that the set of moving points can be effectively learned as an object against the stationary background. Fundamentally, their design principle shares the key spirit of Gestalt theory \cite{Wertheimer1923,Wagemans2012} developed exactly 100 years ago: the raw sensory data with similar motion are likely to be organized into a single object. This is indeed true in the real world where solid objects usually have strong correlation in rigid motions. However, these methods cannot learn to simultaneously segment multiple interested 3D objects from a single point cloud in one go.

Motivated by the potential of motion dynamics, this paper aims to design a general neural framework to simultaneously segment multiple 3D objects, without requiring any human annotations but the inherent object dynamics in training. To achieve this, a na\"ive approach is to train a neural network to directly cluster motion vectors into groups from sequential point clouds, which is widely known as motion segmentation \cite{Yang2021,Yang2019e}. However, such design requires that the input data points are sequential in both training and testing phases, and the trained model cannot infer objects from a single point cloud. Fundamentally, this is because the learned motion segmentation strategies simply cluster similar motion vectors instead of discriminating object geometries, and therefore such design is not general enough for real-world applications. 

\begin{figure}[t]
\centering
  \includegraphics[width=1\linewidth]{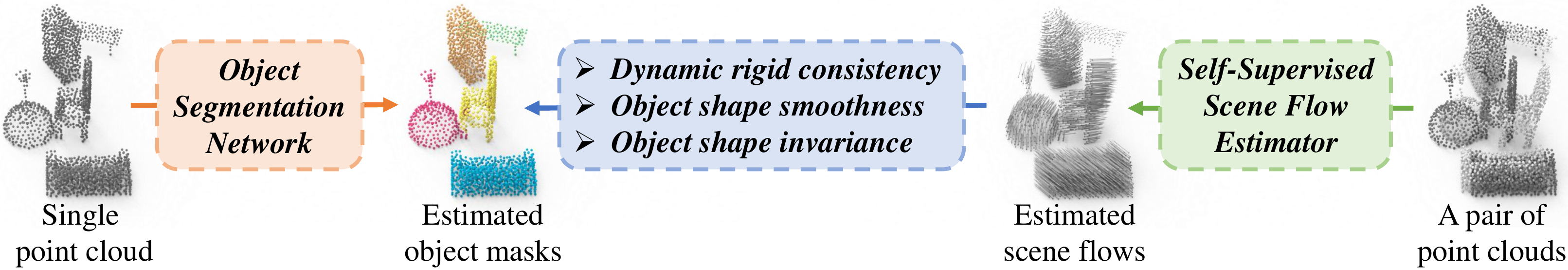}
  \vspace{-0.4cm}
\caption{The general workflow and components of our framework.  }
\label{fig:meth_overview}
\vspace{-0.4cm}
\end{figure}

In this regard, we design a new pipeline which takes a single point cloud as input and directly estimates multiple object masks in a single forward pass. Without needing any human annotations, our pipeline instead leverages the underlying dynamics of sequential point clouds as supervision signals. In particular, as shown in Figure \ref{fig:meth_overview}, our architecture consists of three major components: 1) an object segmentation network to extract per-point features and estimate all object masks from \textit{a single point cloud}, as indicated by the orange block; 2) an auxiliary self-supervised network to estimate per-point motion vectors from a pair of point clouds, as indicated by the green block; 3) a series of loss functions to fully utilize the motion dynamics to supervise the object segmentation backbone, as indicated by the blue block. For the first two components, it is actually flexible to adopt any of existing neural feature extractors \cite{Qi2017} and self-supervised motion estimators \cite{Kittenplon2021}. Nevertheless, the third component is particularly challenging to design, primarily because we need to take into account not only the consistency of diverse dynamics of multiple objects in a sequence, but also the invariance of object geometry irregardless of different moving patterns. 

To tackle this challenge, we introduce three key losses to end-to-end train our object segmentation network from scratch: 1) a multi-object dynamic rigid consistency loss, which aims to evaluate how coherently all estimated object masks (shapes) can fit the motion via rigid transformations; 
2) an object shape smoothness prior, which regularizes all points of each estimated object to be spatially continual instead of fragmented;
3) an object shape invariance loss, which drives multiple estimated masks of a particular object to be invariant given different (augmented) rigid transformations. These losses together force all estimated \textbf{o}bjects' \textbf{g}eometry to be \textbf{c}onsistent and represented by high-quality masks, purely from raw 3D point clouds without any  human annotations. Our method is called \textbf{\nickname{}} and our contributions are:

\begin{itemize}[leftmargin=*]
\setlength{\itemsep}{1pt}
\setlength{\parsep}{1pt}
\setlength{\parskip}{1pt}
  \vspace{-0.4cm}
    \item We introduce the first unsupervised multi-object segmentation pipeline on single point cloud frames, without needing any human annotations in training or multiple frames as input.
    \item We design a set of geometry consistency based losses to fully leverage the object rigid dynamics and shape invariance as effective supervision signals.  
    \item We demonstrate promising object segmentation performance on five datasets, showing significantly better results than classical clustering and optimization baselines.  
      \vspace{-0.4cm}
\end{itemize}

\textbf{Difference from Scene Flow Estimation:} We do not aim to design a new scene flow estimation method such as {\cite{Wu2020,Kittenplon2021,Gojcic2021,Wang2022}}. Instead, we use unsupervised learning based per-point scene flow as supervision signals for single-frame multi-object segmentation.

\textbf{Difference from Motion Segmentation:} We neither aim to segment motion vectors such as \cite{Wang2022,Sun2022} which require multiple successive frames as input in both training and testing. Instead, our network directly estimates object masks from single frames, and therefore is more flexible and general.

\textbf{Scope:} This paper does not intend to replace fully-supervised approaches because the never-moving objects are unlikely to be discovered due to the lack of supervision signals. In addition, estimating object categories or non-rigid objects such as articulated buses and semi-truck with trailers is also out of the scope of this paper.

%% file: chaps/02_liter.tex
\textbf{Fully-supervised 3D Object Segmentation:} To identify 3D objects from point clouds, existing fully-supervised solutions can be divided as 1) bounding box based object detection methods \cite{Zhou2018a,Lang2019,Shi2020} or 2) mask based instance segmentation pipelines \cite{Wang2018d,Yang2019d,Vu2022}. Thanks to the dense human annotations and the well developed backbones including projection-based \cite{Li2016f,Chen2017b,Lang2019}, point-based \cite{Qi2017,Thomas2019,Hu2020} and voxel-based \cite{Graham2018,Choy2019} feature extractors, these methods achieve impressive performance on both indoor and outdoor datasets. However, manually labelling every object in large-scale point clouds is costly. To alleviate this burden, we aim to pioneer 3D object segmentation without human labels. 

\textbf{3D Scene Flow Estimation and Motion Segmentation:} Given sequential point clouds, per-point 3D motion vectors, also known as scene flow, can be accurately estimated. Early works focus on fully-supervised scene flow estimation \cite{Liu2018d,Behl2018,Gu2019,Liu2019i,Puy2020,Ouyang2021,Wei2021}, whereas recent methods start to explore self-supervised motion estimation \cite{Mittal2020,Wu2020,Kittenplon2021,Luo2021,Zeng2021,Li2021}. Taking the scene flow as input, a number of works \cite{Yi2018,Huang2021,Thomas2021,Baur2021} aim to group similar motion vectors, and then obtain bounding boxes or masks only for dynamic objects. Although achieving encouraging results, they either rely on ground truth segmentation for supervision or can only segment simple foreground and background objects, without being able to simultaneously segment multiple objects.
 In this paper, we leverage the successful self-supervised scene flow estimator as our auxiliary neural network to provide valuable supervision signals, so that multiple objects can be identified in a single forward pass. 

\textbf{Unsupervised 2D Object Segmentation:} Inspired by the early work AIR \cite{Eslami2016}, a large number of generative models have been proposed to discover objects from single images without needing human annotations, including MONet \cite{Burgess2019}, IODINE \cite{Greff2019}, Slot-Att \cite{Locatello2020}, \etc{}. These methods are further extended to segment objects from video frames \cite{Kosiorek2018,Hsieh2018,Minderer2019,Jiang2020d,Zoran2021,Ding2021}. However, as investigated by the recent work \cite{Yang2022}, all these approaches can only process simple synthetic datasets, and cannot discover objects from complex real-world images yet. It is still elusive to apply these ideas on 3D point clouds where 3D objects are far more complicated and diverse in terms of geometry. 

\textbf{2D Scene Flow Estimation and Motion Segmentation:} Given image sequences, pixel-level 2D scene flow, also known as optical flow, have been extensively studied in literature \cite{Fortun2015,Zhai2021}. The estimated flow field can be further grouped as objects \cite{Sun2010, Sun2012, Cheng2017,Lu2020,Yang2021,Liu2021c,Kipf2022}. Drawing insights from these works, this paper aims to segment multiple diverse objects in the complex 3D space. 

%% file: chaps/03_meth.tex
\begin{figure}[t]
\centering
  \includegraphics[width=0.95\linewidth]{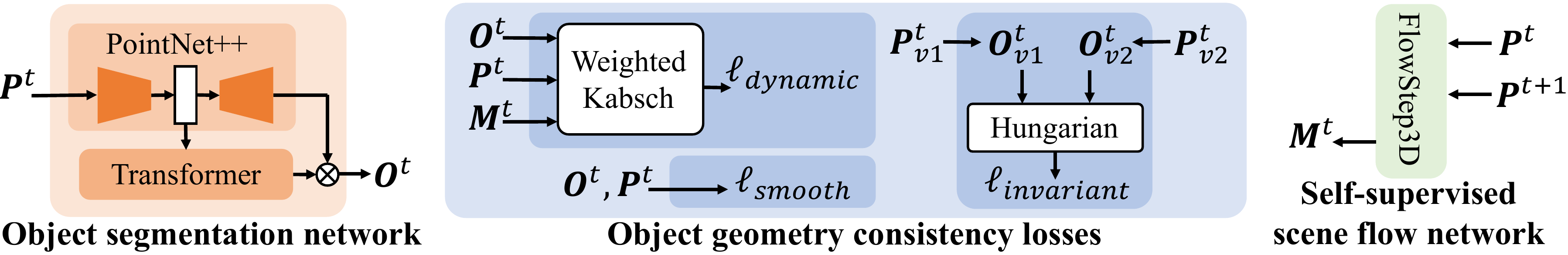}
  \vspace{-0.15cm}
\caption{Components of our pipeline. The object segmentation network consists of PointNet++ and Transformer decoders. FlowStep3D is adopted as the self-supervised scene flow network.}
\label{fig:meth_architecture}
\vspace{-0.2cm}
\end{figure}

\subsection{Overview}
As shown in Figure \ref{fig:meth_architecture}, given a single point cloud $\boldsymbol{P}^t$ with $N$ points as input, \ie{}, $\boldsymbol{P}^t \in \mathbb{R}^{N\times 3}$, where each point only has a location $\{x, y, z\}$ without color for simplicity, the \textbf{object segmentation network} extracts per-point features and directly reasons a set of object masks, denoted as $\boldsymbol{O}^t\in 
\mathbb{R}^{N\times K}$, where $K$ is a predefined number of objects that is large enough for a specific dataset. In particular, we firstly adopt PointNet++ \cite{Qi2017} to extract the per-point local features. Then we employ Transformer decoders \cite{Vaswani2017} to attend to the point features and yield all object masks in parallel. The whole architecture can be regarded as a 3D extension of the recent MaskFormer \cite{Cheng2021} which shows excellent performance in object segmentation in 2D images. Thanks to the powerful Transformer module, each inferred object mask is effectively modeled over the entire point cloud. Implementation details are in Appendix \ref{sec:net_arch}

In the meantime, we have the corresponding sequence of point clouds for supervision, denoted as $\{\boldsymbol{P}^t, \boldsymbol{P}^{t+1}, \cdots\}$. For simplicity, we only use the first two frames $\{\boldsymbol{P}^t, \boldsymbol{P}^{t+1}\}$ and feed them into the \textbf{auxiliary self-supervised scene flow network}, obtaining satisfactory motion vectors for every point in the first point cloud frame, denoted as $\boldsymbol{M}^t \in \mathbb{R}^{N\times 3}$, where each motion vector represents point displacement $\{\Delta x, \Delta y, \Delta z\}$. Among the existing self-supervised scene flow methods, we choose the recent FlowStep3D \cite{Kittenplon2021} which shows excellent scene flow estimation in multiple datasets.
Implementation details are in Appendix \ref{sec:net_arch}. To train the object segmentation network from scratch, the key component is the supervision mechanism as discussed below.

\subsection{Object Geometry Consistency Losses}
Given the input point cloud $\boldsymbol{P}^t$ and its output object masks $\boldsymbol{O}^t$ and motion $\boldsymbol{M}^t$, we introduce the following objectives to satisfy the geometry consistency on both frames $\boldsymbol{P}^t$ and ($\boldsymbol{P}^{t}+\boldsymbol{M}^t$). Note that, the masks $\boldsymbol{O}^t$ are meaningless at the very beginning and need to be optimized appropriately.

\phantom{xxx}\textbf{(1) Geometry Consistency over Dynamic Object Transformations}

From time $t$ to $t+1$, the rigid objects in point cloud frame $\boldsymbol{P}^t$ usually exhibit different dynamic transformations which can be described by matrices belonging to $SE(3)$ group. For the $k^{th}$ object, we firstly retrieve its (soft) binary mask $\boldsymbol{O}^t_k$, and then feed the tuple $\{\boldsymbol{P}^t, \boldsymbol{P}^{t}+\boldsymbol{M}^t, \boldsymbol{O}^t_k\}$ into the differentiable weighted-Kabsch algorithm \cite{kabsch,Gojcic2020}, estimating its transformation matrix $\boldsymbol{T}_k \in \mathbb{R}^{4\times 4}$. 

In order to drive all raw object masks to be more and more accurate, so as to fully explain the corresponding motion patterns within all masks, the following dynamic rigid loss is designed to minimize the discrepancy of per-point scene flow between time $t$ and $t+1$ for each point in $\boldsymbol{P}^t$:
\begin{equation}
    \ell_{dynamic} = \frac{1}{N} \Big\| \Big(\sum_{k=1}^K \boldsymbol{o}_k^t * (\boldsymbol{T}_k \circ \boldsymbol{p}^t) \Big) - (\boldsymbol{p}^t + \boldsymbol{m}^t) \Big\|_2
\end{equation}
where $\boldsymbol{o}_k^t$ and $\boldsymbol{m}^t$ represent the $k^{th}$ object mask and motion of a single point $\boldsymbol{p}^t$ from the point cloud $\boldsymbol{P}^t$, and the operation $\circ$ applys the rigid transformation to that point. Intuitively, if one inferred object mask happens to include two sets of points with two different moving directions, the transformed point cloud can be only in favor of one moving direction, thereby resulting in higher errors.
Therefore, the above constraint can push all object masks to fit the dynamic and diverse motion patterns. However, here arises a critical issue: a single rigid object may be assigned to multiple masks, \ie{} oversegmentation. We alleviate this issue by a simple smoothness regularizer discussed below. 

We observe that such rigid constraint concept is also applied in recent scene flow estimation method \cite{Gojcic2021}. However, their objective is to push the scene flow to be consistent given object masks (estimated by DBSCAN clustering), while our objective is to learn high-quality masks from given flows.

\phantom{xxx}\textbf{(2) Geometry Smoothness Regularization} 

The primary reason why a single object may be oversegmented is the lack of spatial connectivity between individual points. However, our common observation is that physically neighbouring points usually belong to a single object. In this regard, we simply introduce a geometry smoothness regularizer. Particularly, for a specific $n^{th}$ 3D point $\boldsymbol{p}_n$ in the point cloud $\boldsymbol{P}^t$, we firstly search $H$ points from its neighbourhood using either KNN or spherical querying methods, and then force their mask assignments to be consistent with the center point $\boldsymbol{p}_n$. Mathmatically, it is defined as:
\begin{equation}
    \ell_{smooth} = \frac{1}{N} \sum_{n=1}^N \Big( \frac{1}{H}\sum_{h=1}^H d( \boldsymbol{o}_{p_n}, \boldsymbol{o}_{p_n^h} ) \Big)
\end{equation}
where $\boldsymbol{o}_{p_n} \in \mathbb{R}^{1\times K}$ represents the object assignment of center point $\boldsymbol{p}_n$, and $\boldsymbol{o}_{p_n^h} \in \mathbb{R}^{1\times K}$ represents the object assignment of its $h^{th}$ neighbouring point. The distance function $d()$ is flexible to choose $L1$ / $L2$ or a more aggressive cross-entropy function.

Note that, such local smoothness prior is successfully used for scene flow estimation \cite{Liu2018d,Kittenplon2021}. Here, we instead demonstrate its effectiveness for object segmentation.


\phantom{xxx}\textbf{(3) Geometry Invariance over Scene Transformations}

With the above geometry constraints designed in (1)(2), the shapes of dynamic objects can be reasonably segmented. However, the learned object geometry may not be general enough. For example, a moving car can be well segmented, yet another similar parked car may not be discovered. 
To this end, we introduce an object geometry invariance constraint as follows: 
\begin{itemize}[leftmargin=*]
\setlength{\itemsep}{1pt}
\setlength{\parsep}{1pt}
\setlength{\parskip}{1pt}
\item Firstly, given $\boldsymbol{P}^t$, we apply two transformations to get augmented point clouds $\boldsymbol{P}^t_{v1}$ and $\boldsymbol{P}^t_{v2}$.
\item Secondly, we feed $\boldsymbol{P}^t_{v1}$ and $\boldsymbol{P}^t_{v2}$ into our object segmentation network, obtaining two sets of object masks $\boldsymbol{O}^t_{v1}$ and $\boldsymbol{O}^t_{v2}$. Because the per-point locations in two point clouds are transformed differently, the position sensitive PointNet++ \cite{Qi2017} features generate two different sets of masks.
\item Thirdly, we leverage the Hungarian algorithm \cite{Kuhn1955} to one-one match the individual masks in $\boldsymbol{O}^t_{v1}$ and $\boldsymbol{O}^t_{v2}$ according to the object pair-wise IoU scores. Basically, this is to address the issue that there is no fixed order for predicted object masks from the two augmented point clouds.
\item At last, we reorder the masks in $\boldsymbol{O}^t_{v2}$ to align with $\boldsymbol{O}^t_{v1}$, and design the invariance loss as follows. 
\end{itemize}\vspace{-4pt}
\begin{equation}
    \ell_{invariant} = \frac{1}{N}\sum_{n=1}^{N}\hat{d}\big(\boldsymbol{\hat{o}}_{v1}^{n}, \boldsymbol{\hat{o}}_{v2}^{n} \big)
\end{equation}
where $\boldsymbol{\hat{o}}_{v1}^n$ and $\boldsymbol{\hat{o}}_{v2}^n$ are the reordered object assignments of the two augmented point clouds for a specific $n^{th}$ point. The distance function $\hat{d}()$ is flexible to use $L1$, $L2$ or cross-entropy. Ultimately, this loss drives the estimated object masks to be invariant with different views of input point clouds. 

Notably, unlike existing self-supervised learning \cite{Cho2021} which usually uses invariance prior for better latent representations, here we aim to generalize the segmentation strategy to similar yet static objects.

\subsection{Iterative Optimization of Object Segmentation and Motion Estimation}
With the designed geometry consistency loss functions, the object segmentation network is optimized from scratch by the combined loss: $\ell_{seg} = \ell_{dynamic} + \ell_{smooth} + \ell_{invariant}$. For efficiency, the auxiliary self-supervised scene flow network FlowStep3D \cite{Kittenplon2021} is independently trained by its own losses until convergence. Intuitively, with better and better object masks estimated, the estimated scene flow is also expected to be improved further if we use the masks properly. To this end, we propose the following Algorithm \ref{alg:iterative} to iteratively improve object segmentation and motion estimation. 

\vspace{-0.2cm}
\begin{algorithm}[H]
\caption{Iterative optimization of object segmentation and scene flow estimation. Assume the whole train split has $S$ point cloud pairs: $\{(\boldsymbol{P}^t, \boldsymbol{P}^{t+1})_1 \cdots  (\boldsymbol{P}^t, \boldsymbol{P}^{t+1})_S\}$.
}
\label{alg:iterative}
\begin{algorithmic} 
\STATE{\textit{Stage 0: Initial scene flow estimation.}} 
\STATE{\phantom{xx}$\bullet$} Independently and fully train the self-supervised scene flow network on the whole training data split, and obtain reasonable scene flow estimations: $\{(\boldsymbol{P}^t, \boldsymbol{P}^{t+1}, \boldsymbol{M}^t)_1 \cdots  (\boldsymbol{P}^t, \boldsymbol{P}^{t+1}, \boldsymbol{M}^t)_S\}$.

\FOR {number of iteration rounds $R$}{}
\STATE{\textit{Stage 1: Object segmentation optimization.}} 

\STATE{\phantom{xx}$\bullet$ Train the object segmentation network using $\ell_{seg}$ for a total $E$ epochs on the whole training split: $\{(\boldsymbol{P}^t, \boldsymbol{P}^{t+1}, \boldsymbol{M}^t)_1 \cdots  (\boldsymbol{P}^t, \boldsymbol{P}^{t+1}, \boldsymbol{M}^t)_S\}$.
}

\STATE{\phantom{xx}$\bullet$ Estimate reasonable object masks: $\{(\boldsymbol{P}^t, \boldsymbol{P}^{t+1}, \boldsymbol{O}^t, \boldsymbol{O}^{t+1})_1 \cdots  (\boldsymbol{P}^t, \boldsymbol{P}^{t+1},s \boldsymbol{O}^t, \boldsymbol{O}^{t+1})_S\}$.}

\STATE{\textit{Stage 2: Scene flow improvement.}}
\STATE{\phantom{xx}$\bullet$ For each pair of data $(\boldsymbol{P}^t, \boldsymbol{P}^{t+1}, \boldsymbol{M}^t, \boldsymbol{O}^t, \boldsymbol{O}^{t+1})$, by drawing insights from the classical ICP \cite{Arun1987}, we propose an \textbf{object-aware ICP} algorithm to estimate new scene flow $\boldsymbol{\hat{M}}^t$ for point cloud $\boldsymbol{P}^t$.     
}
\STATE{\phantom{xx}$\bullet$ Update the new scene flow for next round training:\\ 
$\{(\boldsymbol{P}^t, \boldsymbol{P}^{t+1}, \boldsymbol{M}^t)_1 \cdots  (\boldsymbol{P}^t, \boldsymbol{P}^{t+1}, \boldsymbol{M}^t)_S\} \leftarrow \{(\boldsymbol{P}^t, \boldsymbol{P}^{t+1}, \boldsymbol{\hat{M}}^t)_1 \cdots  (\boldsymbol{P}^t, \boldsymbol{P}^{t+1}, \boldsymbol{\hat{M}}^t)_S\}$
}
\ENDFOR
\end{algorithmic}
\end{algorithm}
\vspace{-0.5cm}

Empirically, setting the total number of rounds $R$ to be 2 or 3 has a good trade off between accuracy and training efficiency. Due to the limited space, details of object-aware ICP algorithm are in Appendix \ref{sec:oa_icp}. We exclude the invariance loss $l_{invariant}$ from object segmentation optimization stage in the early rounds so that the networks can focus on moving objects in training and produce better scene flows, and then add $l_{invariant}$ back in the final round. Detailed analysis is in Appendix \ref{sec:more_abla}.



%% file: chaps/04_exp.tex
Our method is evaluated on four different application scenarios: 1) part instance segmentation of articulated objects on SAPIEN dataset \cite{Xiang2020}, 2) object segmentation of indoor scenes on our own synthetic dataset, 3) object segmentation of real-world outdoor scenes on KITTI-SF dataset \cite{Menze2015}, and 4) object segmentation on the sparse yet large-scale LiDAR based KITTI-Det \cite{Geiger2012} and SemanticKITTI \cite{Behley2019} datasets. For evaluation metrics, we follow \cite{Nunes2018} and report the \textbf{F1-score}, \textbf{Precision}, and \textbf{Recall} with an IoU threshold of 0.5. In addition, we report the Average Precision (\textbf{AP}) score following COCO dataset \cite{Lin2014} and the Panoptic Quality (\textbf{PQ}) score defined in \cite{Kirillov2019}. The mean Intersection over Union (\textbf{mIoU}) score and the Rand Index (\textbf{RI}) score implemented in \cite{Huang2021} are also included. Note that, all metrics  are computed in a class-agnostic manner.

\subsection{Evaluation on SAPIEN Dataset}
\label{sec:eval_sapien}

The SAPIEN dataset \cite{Xiang2020} provides 720 simulated articulated objects with part instance level annotations. Each object has 4 sequential scans. The part instances have different articulating (moving) states. We follow \cite{Huang2021} to use the training data generated from \cite{Yi2016}. In particular, there are 82092 pairs of point clouds for training, 2880 single point cloud frames for testing. Each point cloud is downsampled to 512 points in both training and testing. 

Since there is no existing unsupervised method for multi-object segmentation on 3D point clouds, we firstly implement two classical clustering methods: WardLinkage \cite{Jr.1963} and DBSCAN \cite{Ester1996} to directly group 3D points from single point clouds into objects. Secondly, we implement two classical motion segmentation methods: TrajAffn \cite{Ochs2014} and SSC \cite{Nunes2018}. Note that, these two methods take the same estimated scene flows of FlowStep3D as input, while our method uses the estimated scene flows during training only, but takes single point clouds as input during testing. In addition, we also include the excellent results of several fully-supervised methods (PointNet++ \cite{Qi2017}, MeteorNet \cite{Liu2019i}, DeepPart \cite{Yi2018}) reported in MBS \cite{Huang2021}. Their experimental details can be found in MBS \cite{Huang2021}. Lastly, we train our object segmentation network using single point clouds with full annotations, denoted as \nickname{}$_{sup}$. All implementation details are in Appendix \ref{sec:more_implement}. 

\textbf{Analysis:} As shown in Table \ref{tab:res_sapien}, our \nickname{} surpasses the classical clustering based and motion segmentation methods by large margins on all metrics, showing the advantage of our method in fully leveraging both the motion patterns and various types of geometry consistency. Compared with the fully supervised baselines, our method is only inferior to the strong MBS \cite{Huang2021} and \nickname{}$_{sup}$. However, we observe that our \nickname{} actually shows a higher precision score than \nickname{}$_{sup}$, primarily because our method tends to learn better objectness thanks to a combination of motion pattern and smoothness constraints and avoid dividing a single object into pieces. Figure \ref{fig:qual_result} shows qualitative results.

\begin{table}
      \tabcolsep= 0.19cm 
  \caption{Quantitative results of our method and baselines on the SAPIEN dataset.}
  \label{tab:res_sapien}
  \centering
  \begin{tabular}{crccccccc}
    \toprule
    & & AP$\uparrow$ & PQ$\uparrow$ & F1$\uparrow$ & Pre$\uparrow$ & Rec$\uparrow$ & mIoU$\uparrow$ & RI$\uparrow$ \\ 
    \midrule
    \multirow{5}{*}{\makecell[c]{Supervised\\Methods} }& PointNet++ \cite{Qi2017} & - & - & - &- &- & 51.2 & 65.0 \\
    & MeteorNet \cite{Liu2019i} & - &- &- &- &- & 45.7 & 60.0 \\
    & DeepPart \cite{Yi2018} &- &- &- &- &- & 53.0 & 67.0 \\
    & MBS \cite{Huang2021} &- &- &- &- &- & 67.3 & 77.0 \\
    & \nickname{}$_{sup}$ & 66.1 & 48.7 & 62.0 & 54.6 & 71.7 & 66.8 & 77.1 \\  
    \midrule
    \multirow{2}{*}{\makecell[c]{Unsupervised\\Motion Segmentation}}& TrajAffn \cite{Ochs2014} & 6.2 & 14.7 & 22.0 & 16.3 & 34.0 & 45.7 & 60.1 \\
    & SSC \cite{Nunes2018} & 9.5 & 20.4 & 28.2 & 20.9 & 43.5 & 50.6 & 65.9 \\ 
    \hdashline
    \multirow{3}{*}{\makecell[c]{Unsupervised\\Methods}}& WardLinkage \cite{Jr.1963} & 17.4 & 26.8 & 40.1 & 36.9 & 43.9 & 49.4 & 62.2 \\
    & DBSCAN \cite{Ester1996} & 6.3 & 13.4 & 20.4 & 13.9 & 37.9 & 34.2 & 51.4 \\
    & \textbf{\nickname{}(Ours)} & \textbf{55.6} & \textbf{50.6} & \textbf{65.1} & \textbf{65.0} & \textbf{65.2} & \textbf{60.9} & \textbf{73.4} \\  
    \bottomrule
  \end{tabular}
  \vspace{-0.4cm}
\end{table}

\subsection{Evaluation on \nickname{}-DR / \nickname{}-DRSV Datasets}
\label{sec:eval_indoor}

We further evaluate our method to segment objects in indoor 3D scenes. Considering that the existing dataset FlyingThings3D \cite{Mayer2016} tends to have unrealistically cluttered scenes with severely fragmented objects and it is originally introduced for scene flow estimation, we turn to synthesize a new dynamic room dataset, called \textbf{\nickname{}-DR}, that suits both scene flow estimation and object segmentation. In particular, we follow \cite{Peng2020a} to randomly place $4 \sim 8$ objects belonging to 7 classes of ShapeNet \cite{Chang2015} \{chair, table, lamp, sofa, cabinet, bench, display\} into each room. In total, we create 3750, 250, and 1000 indoor rooms (scenes) for training/validation/test splits. In each scene, we create rigid dynamics by applying continuous random transformations to each object and record 4 sequential frames for evaluation. Each point cloud frame is downsampled to 2048 points. Note that, we follow \cite{Chan2016} to split different object instances for train/val/test sets. 

Based on our \nickname{}-DR dataset, we collect single depth scans every time step on the mesh models to generate another dataset, called Single-View \nickname{}-DR (\textbf{\nickname{}-DRSV}). All object point clouds in \nickname{}-DRSV are severely incomplete due to self- and/or mutual occlusions, resulting in the new dataset significantly more challenging than \nickname{}-DR. Each point cloud frame in \nickname{}-DRSV is also downsampled to 2048 points. More details of these two datasets are in Appendix \ref{sec:ogcdr_datagen}.

\textbf{Analysis:} As shown in Table \ref{tab:res_indoor}, our method outperforms all classical unsupervised methods including the clustering based and the motion segmentation based methods on \nickname{}-DR. Since the synthetic rooms in \nickname{}-DR all have complete 3D objects, and the generated point cloud sequences are of high quality. Therefore, our \nickname{} even surpasses the supervised \nickname{}$_{sup}$. This shows that the rigid dynamic motions can indeed provide sufficient supervision signals to identify objects. On \nickname{}-DRSV, our method still achieves superior performance and demonstrates robustness to incomplete point clouds, although the scores are slightly lower than that on the full point cloud dataset \nickname{}-DR (AP: 86.8 $vs$ 92.3).  Figure \ref{fig:qual_result} shows qualitative results.

\begin{table}[t]
      \tabcolsep= 0.145cm 
  \caption{Quantitative results of our method and baselines on our \nickname{}-DR/\nickname{}-DRSV dataset.}
  \label{tab:res_indoor}
  \centering
  \scriptsize
  \begin{tabular}{crccccccc}
    \toprule
     & & AP$\uparrow$ & PQ$\uparrow$ & F1$\uparrow$ & Pre$\uparrow$ & Rec$\uparrow$ & mIoU$\uparrow$ & RI$\uparrow$ \\
     \midrule
     Supervised Method & \nickname{}$_{sup}$ & 90.7 / 86.3 & 82.6 / 78.8 & 87.6 / 85.0 & 83.7 / 82.2 & 92.0 / 88.0 & 89.2 / 83.9 & 97.7 / 97.1 \\
     \midrule
     \multirow{2}{*}{\makecell[c]{Unsupervised\\Motion Segmentation}} & TrajAffn \cite{Ochs2014} & 42.6 / 39.3 & 46.7 / 43.8 & 57.8 / 54.8 & 69.6 / 63.0 & 49.4 / 48.4 & 46.8 / 45.9 & 80.1 / 77.7 \\
     & SSC \cite{Nunes2018} & 74.5 / 70.3 & 79.2 / 75.4 & 84.2 / 81.5 & 92.5 / 89.6 & 77.3 / 74.7 & 74.6 / 70.8 & 91.5 / 91.3 \\  
     \hdashline
     \multirow{3}{*}{\makecell[c]{Unsupervised\\Methods}} & WardLinkage \cite{Jr.1963} & 72.3 / 69.8 & 74.0 / 71.6 & 82.5 / 80.5 & \textbf{93.9 / 91.8} & 73.6 / 71.7 & 69.9 / 67.2 & 94.3 / 93.3 \\
     & DBSCAN \cite{Ester1996} & 73.9 / 71.9 & 76.0 / 76.3 & 81.6 / 81.8 & 85.8 / 79.1 & 77.8 / 84.8 & 74.7 / 80.1 & 91.5 / 93.5 \\
     & \textbf{\nickname{}(Ours)} & \textbf{92.3 / 86.8} & \textbf{85.1 / 77.0} & \textbf{89.4 / 83.9} & 85.6 / 77.7 & \textbf{93.6 / 91.2} & \textbf{90.8 / 84.8} & \textbf{97.8 / 95.4} \\
    \bottomrule
  \end{tabular}
    \vspace{-0.5cm}
\end{table}

\subsection{Evaluation on KITTI Scene Flow Dataset}
\label{sec:eval_kittisf}

We additionally evaluate our method on the challenging real-world outdoor KITTI Scene Flow (KITTI-SF) dataset. Officially, KITTI-SF dataset \cite{Menze2015} consists of 200 (training) pairs of point clouds from real-world traffic scenes and an online hidden test for scene flow estimation. In our experiment, we train our pipeline on the first 100 pairs of point clouds, and then test on the remaining 100 pairs (200 single point clouds). We observe that in the 100 training pairs, the moving objects are only cars and trucks. Therefore, in the testing phase, we only keep the human annotations \cite{Alhaija2018} of cars and trucks in every single frame to compute the scores. All other objects are treated as part of background. Note that, the whole background is not ignored, but counted as one object in our evaluation, and the cars and trucks can be static or moving. We find KITTI-SF is too challeging for the classical unsupervised methods, due to the extreme imbalance of 3D points between objects and background. Besides, the background and objects in KITTI-SF are always connected because of the Earth's gravity, while clustering-based WardLinkage and DBSCAN favor spatially separated objects. Therefore, we leverage the prior about ground planes in KITTI-SF to assist these methods. We detect and specially handle the ground planes, leaving above-ground points only for these methods to handle. Implementation details are in Appendix \ref{sec:more_implement}.

\textbf{Analysis:} As shown in Table \ref{tab:res_kittisf}, our method obtains superior segmentation scores on the KITTI-SF dataset, being very close to our fully-supervised counterpart \nickname{}$_{sup}$. This demonstrates the excellence of our method on real-world scenes. Figure \ref{fig:qual_result} shows qualitative results.

\vspace{-0.2cm}
\begin{table}[h]
      \tabcolsep= 0.19cm 
  \caption{Quantitative results of our method and baselines on the KITTI-SF dataset.}
  \label{tab:res_kittisf}
  \centering
  \begin{tabular}{crccccccc}
    \toprule
    & & AP$\uparrow$ & PQ$\uparrow$ & F1$\uparrow$ & Pre$\uparrow$ & Rec$\uparrow$ & mIoU$\uparrow$ & RI$\uparrow$ \\
    \midrule
    Supervised Method & \nickname{}$_{sup}$ & 62.4 & 52.7 & 65.1 & 63.4 & 67.0 & 67.3 & 95.0 \\
    \midrule
    \multirow{2}{*}{\makecell[c]{Unsupervised\\Motion Segmentation}} & TrajAffn \cite{Ochs2014} & 24.0 & 30.2 & 43.2 & 37.6 & 50.8 & 48.1 & 58.5 \\
    & SSC \cite{Nunes2018} & 12.5 & 20.4 & 28.4 & 22.8 & 37.6 & 41.5 & 48.9 \\ 
    \hdashline
    \multirow{3}{*}{\makecell[c]{Unsupervised\\Methods}} & WardLinkage \cite{Jr.1963} & 25.0 & 16.3 & 22.9 & 13.7 & \textbf{69.8} & 60.5 & 44.9 \\
    & DBSCAN \cite{Ester1996} & 13.4 & 22.8 & 32.6 & 26.7 & 42.0 & 42.6 & 55.3 \\
    & \textbf{\nickname{}(Ours)} & \textbf{54.4} & \textbf{42.4} & \textbf{52.4} & \textbf{47.3} & 58.8 & \textbf{63.7} & \textbf{93.6} \\
    \bottomrule
  \end{tabular}
  \vspace{-0.4cm}
\end{table}

\subsection{Generalization to KITTI Detection and SemanticKITTI Datasets}
\label{sec:eval_kittidet}

Given our well trained model on KITTI-SF in Section \ref{sec:eval_kittisf}, we directly test it on the popular KITTI 3D Object Detection (KITTI-Det) \cite{Geiger2012} and SemanticKITTI \cite{Behley2019} benchmarks. Unlike the stereo-based point clouds in KITTI-SF, point clouds in these two datasets are collected by LiDAR sensors and thus more sparse.

\begin{itemize}[leftmargin=*]
\setlength{\itemsep}{1pt}
\setlength{\parsep}{1pt}
\setlength{\parskip}{1pt}

\item \textbf{KITTI-Det} officially has 3712 point cloud frames for training, 3769 for validation. We only keep the ground truth object masks obtained from bounding boxes for the car category in each frame. All other objects are treated as part of background. For comparison, we download the official pretrained models of three fully-supervised methods PointRCNN \cite{Shi2019}, PV-RCNN \cite{Shi2019a} and Voxel-RCNN \cite{Deng2021} to directly test on the validation split using the same settings as ours. In addition, we use the well trained \nickname{}$_{sup}$ on KITTI-SF to directly test for comparison, denoted as \nickname{}*$_{sup}$. We also train \nickname{}$_{sup}$ on the training split (3712 frames) from scratch and test it on the remaining 3769 frames using the same evaluation settings, denoted as \nickname{}$_{sup}$.

\item \textbf{SemanticKITTI} officially has 11 sequences with annotations for training and another 11 sequences for online hidden test. We only keep ground truth objects of car and truck categories. The total 11 training sequences (23201 point cloud frames) are used for testing. Compared with KITTI-Det, SemanticKITTI holds $6\times$ more testing frames and covers more diverse scenes. Following the official split in \cite{Behley2019}, we also report the results on: i) sequences 00$\sim$07 and 09$\sim$10 (19130 frames), and ii) the sequence 08 (4071 frames) separately.
\end{itemize}

\begin{table}[h]
  \vspace{-0.2cm}
      \tabcolsep= 0.19cm 
  \caption{Quantitative results on KITTI-Det (* denotes the model trained on KITTI-SF).}
  \label{tab:res_kittidet}
  \centering
  \begin{tabular}{crccccccc}
    \toprule
    & & AP$\uparrow$ & PQ$\uparrow$ & F1$\uparrow$ & Pre$\uparrow$ & Rec$\uparrow$ & mIoU$\uparrow$ & RI$\uparrow$ \\
    \midrule
    \multirow{5}{*}{\makecell[c]{Supervised\\Methods}} & PointRCNN \cite{Shi2019} & 95.7 & 80.1 & 88.9 & 81.3 & 98.0 & 91.4 & 97.2 \\
    & PV-RCNN \cite{Shi2019a} & 95.4 & 77.3 & 84.4 & 73.7 & 98.8 & 92.7 & 97.1 \\
    & Voxel-RCNN \cite{Deng2021} & 95.8 & 79.6 & 87.3 & 78.1 & 98.9 & 92.6 & 97.3 \\
    & \nickname{}$_{sup}$ & 80.0 & 68.5 & 78.3 & 72.7 & 84.8 & 84.0 & 96.9 \\
    & \nickname{}*$_{sup}$ & 51.4 & 41.0 & 49.1 & 43.7 & 56.0 & 66.2 & 91.0 \\
     \midrule
    Unsupervised Method & \textbf{\nickname{}*(Ours)}& 40.5 & 30.9 & 37.0 & 30.8 & 46.5 & 60.6 & 86.4 \\
    \bottomrule
  \end{tabular}
  \vspace{-0.2cm}
\end{table}

\setlength{\abovecaptionskip}{-10 pt}
\begin{wraptable}{r}{9.4cm}
  \tabcolsep= 0.06cm
  \caption{Quantitative results on SemanticKITTI (* denotes the model trained on KITTI-SF).}
  \label{tab:res_semkitti}
  \centering
  \begin{tabular}{crccccccc}
    \toprule
    Sequences & Methods & AP$\uparrow$ & PQ$\uparrow$ & F1$\uparrow$ & Pre$\uparrow$ & Rec$\uparrow$ & mIoU$\uparrow$ & RI$\uparrow$ \\
    \midrule
    \multirow{2}{*}{\makecell[c]{00$\sim$10}} & \nickname{}*$_{sup}$ & 53.8 & 41.3 & 48.1 & 40.1 & 60.0 & 68.3 & 90.0 \\
    & \textbf{\nickname{}*(Ours)} & 42.6 & 30.2 & 35.3 & 28.2 & 47.3 & 60.3 & 86.0 \\
    \hline
    \multirow{2}{*}{\makecell[c]{00$\sim$07 \&\\09$\sim$10}} & \nickname{}*$_{sup}$ & 55.3 & 41.8 & 48.4 & 40.1 & 61.1 & 69.9 & 90.3 \\
    & \textbf{\nickname{}*(Ours)} & 43.6 & 30.5 & 35.5 & 28.1 & 48.2 & 62.1 & 86.3 \\
    \hdashline
    \multirow{2}{*}{\makecell[c]{08}} & \nickname{}*$_{sup}$ & 49.4 & 39.2 & 46.6 & 40.0 & 55.8 & 60.3 & 88.3 \\
    & \textbf{\nickname{}*(Ours)} & 38.6 & 29.1 & 34.7 & 28.6 & 44.0 & 51.8 & 84.3 \\
    \bottomrule
  \end{tabular}
    \vspace{-0.4cm}
\end{wraptable}

\textbf{Analysis:} As shown in Tables \ref{tab:res_kittidet}\&\ref{tab:res_semkitti}, our method can directly generalize to 3D object segmentation on sparse LiDAR point clouds with satisfactory results, also being close to the fully-supervised counterpart \nickname{}*$_{sup}$. It is understandable that the other three fully-supervised models have a clear advantage over ours on KITTI-Det, because they are fully supervised and trained on the KITTI-Det training split (3712 frames) while ours does not. We hope that our method can serve the first baseline and inspire more advanced unsupervised methods in the future to close the gap. Figure \ref{fig:qual_result} shows qualitative results.

\subsection{Ablation Study}

\setlength{\abovecaptionskip}{ -10 pt}
\begin{wraptable}{r}{8cm}
  \tabcolsep= 0.07cm 
  \caption{Ablation studies about loss designs on SAPIEN.}
  \label{tab:abl_sapien}
  \centering
  \begin{tabular}{rccccccc}
    \toprule
    & AP$\uparrow$ & PQ$\uparrow$ & F1$\uparrow$ & Pre$\uparrow$ & Rec$\uparrow$ & mIoU$\uparrow$ & RI$\uparrow$ \\ \midrule
    w/o $\ell_{dynamic}$ & 35.4 & 35.3 & 54.1 & \textbf{91.1} & 38.5 & 28.6 & 52.7 \\
    w/o $\ell_{smooth}$ & 21.8 & 18.5 & 26.9 & 19.1 & 45.4 & 52.4 & 63.7 \\
    w/o $\ell_{invariant}$ & 48.9 & 46.1 & 61.3 & 61.9 & 60.7 & 57.9 & 70.3 \\
    Full \nickname{} & \textbf{55.6} & \textbf{50.6} & \textbf{65.1} & 65.0 & \textbf{65.2} & \textbf{60.9} & \textbf{73.4} \\
    \bottomrule
  \end{tabular}
\vspace{-0.3 cm}
\end{wraptable}

\textbf{(1) Geometry Consistency Losses:} To validate the choice of our design, we firstly conduct three groups of ablative experiments on the SAPIEN dataset \cite{Xiang2020}: 1) only remove the dynamic rigid loss $\ell_{dynamic}$, 2) only remove the smoothness loss $\ell_{smooth}$, and 3) only remove the invariance loss $\ell_{invariant}$. As shown in Table \ref{tab:abl_sapien}, combining the proposed three losses together gives the highest segmentation scores. Basically, the dynamic rigid loss serves to discriminate multiple objects from different motion patterns. Without it, the network tends to assign all points to a single object as the shortcut to minimize the other two losses. However, we observe in Table \ref{tab:abl_sapien} that without $\ell_{dynamic}$, the network still works to some extend. This is because the synthetic SAPIEN dataset tends to have a number of point cloud frames with only 2 or 3 objects, thus assigning all points to a single object can still get plausible scores. This issue is further validated by conducting additional ablation experiments on curated SAPIEN dataset. More details are in Appendix \ref{sec:more_abla}.  

In addition, we evaluate the robustness of our object segmentation method with regard to different types of motion estimations, and different hyperparameter and design choices of our smoothness loss $\ell_{smooth}$. More results are in Appendix \ref{sec:more_abla}. 

\setlength{\abovecaptionskip}{-10 pt}
\begin{wraptable}{r}{8cm}
  \tabcolsep= 0.16cm 
  \caption{Iterative optimization on SAPIEN.}
  \label{tab:iter_sapien}
  \centering
  \begin{tabular}{cccccccl} \toprule
    & \multicolumn{7}{c}{Object Segmentation}  \\
    \#$R$ & AP$\uparrow$ & PQ$\uparrow$ & F1$\uparrow$ & Pre$\uparrow$ & Rec$\uparrow$ & mIoU$\uparrow$ & RI$\uparrow$ \\ \midrule
    1 & 45.9 & 47.7 & 62.3 & 60.2 & 64.5 & 60.2 & 72.3 \\
    2 & 55.6 & 50.6 & 65.1 & 65.0 & 65.2 & 60.9 & 73.4 \\
    3 & 56.3 & 50.7 & 65.4 & 65.1 & 65.8 & 61.1 & 73.7 \\ \bottomrule
  \end{tabular}
\vspace{-0.3 cm}
\end{wraptable}

\textbf{(2) Iterative Optimization Algorithm:} We also conduct ablative experiments to validate the effectiveness of our proposed Algorithm \ref{alg:iterative}. We set the number of iterative rounds $R$ as $\{1, 2, 3\}$. As shown in Table \ref{tab:iter_sapien}, after 2 rounds, satisfactory segmentation results can be achieved, although we expect better results after more rounds with longer training time. This shows that our iterative optimization algorithm can indeed fully leverage the mutual benefits between object segmentation and motion estimation. 

\setlength{\abovecaptionskip}{-10 pt}
\begin{wraptable}{r}{8cm}
  \tabcolsep= 0.08cm 
  \caption{Scene flow estimation on the KITTI-SF dataset.}
  \label{tab:res_sceneflow}
  \centering
  \begin{tabular}{rllll}
    \toprule
    & EPE3D$\downarrow$ & AccS$\uparrow$ & AccR$\uparrow$ & Outlier$\downarrow$ \\
    \midrule
    Ego-motion~\cite{Tishchenko20} & 41.54 & 22.09 & 37.21 & 80.96 \\
    PointPWC-Net~\cite{Wu2020} & 25.49 & 23.79 & 49.57 & 68.63 \\
    FlowStep3D~\cite{Kittenplon2021} & 10.21 & 70.80 & 83.94 & 24.56 \\
    \textbf{\nickname{}(Ours)} & \textbf{6.72} & \textbf{80.16} & \textbf{89.08} & \textbf{22.56} \\
    \bottomrule
  \end{tabular}
\vspace{-0.3 cm}
\end{wraptable}

\subsection{Pushing the Boundaries of Unsupervised Scene Flow Estimation and Segmentation}
In addition to the improvement of object segmentation from our iterative optimization algorithm, the scene flow estimation can be naturally further improved from our estimated object masks as well. Given our well trained model on the KITTI-SF dataset in Section \ref{sec:eval_kittisf}, we use the estimated object masks to further improve the scene flow estimation. As shown in Table \ref{tab:res_sceneflow}, following the exact evaluation settings of FlowStep3D \cite{Kittenplon2021}, our method, not surprisingly, significantly boosts the scene flow accuracy, surpassing the state-of-the-art unsupervised FlowStep3D \cite{Kittenplon2021} and other baselines in all metrics. 

\setlength{\abovecaptionskip}{ -10 pt}
\begin{wraptable}{r}{8cm}
  \tabcolsep= 0.1cm 
  \caption{Motion $vs$ Points based segmentation on \tiny{KITTI-SF}.}
  \label{tab:res_motionseg}
  \centering
  \begin{tabular}{rccccccc} 
    \toprule
    Input & AP$\uparrow$ & PQ$\uparrow$ & F1$\uparrow$ & Pre$\uparrow$ & Rec$\uparrow$ & mIoU$\uparrow$ & RI$\uparrow$ \\ 
    \midrule
    scene flow & 47.3 & 41.2 & 50.2 & \textbf{50.9} & 49.6 & 56.0 & 89.3 \\
    point cloud & \textbf{54.4} & \textbf{42.4} & \textbf{52.4} & 47.3 & \textbf{58.8} & \textbf{63.7} & \textbf{93.6} \\
    \bottomrule
  \end{tabular}
\vspace{-0.3 cm}
\end{wraptable}
In fact, our object segmentation backbone is also flexible to take the scene flow as input instead of point $xyz$ to segment objects. This is commonly called motion segmentation. We replace the single point clouds by (estimated) scene flow vectors as our network inputs, and train the network from scratch using the same settings on the KITTI-SF dataset. As shown in Table \ref{tab:res_motionseg}, we can see that our network can still achieve superior results regardless of the modality of inputs, demonstrating the generality of our framework.

%% file: chaps/05_sum.tex
\begin{figure}[t]
\setlength{\abovecaptionskip}{ 4 pt}
\centering
  \includegraphics[width=1.0\linewidth]{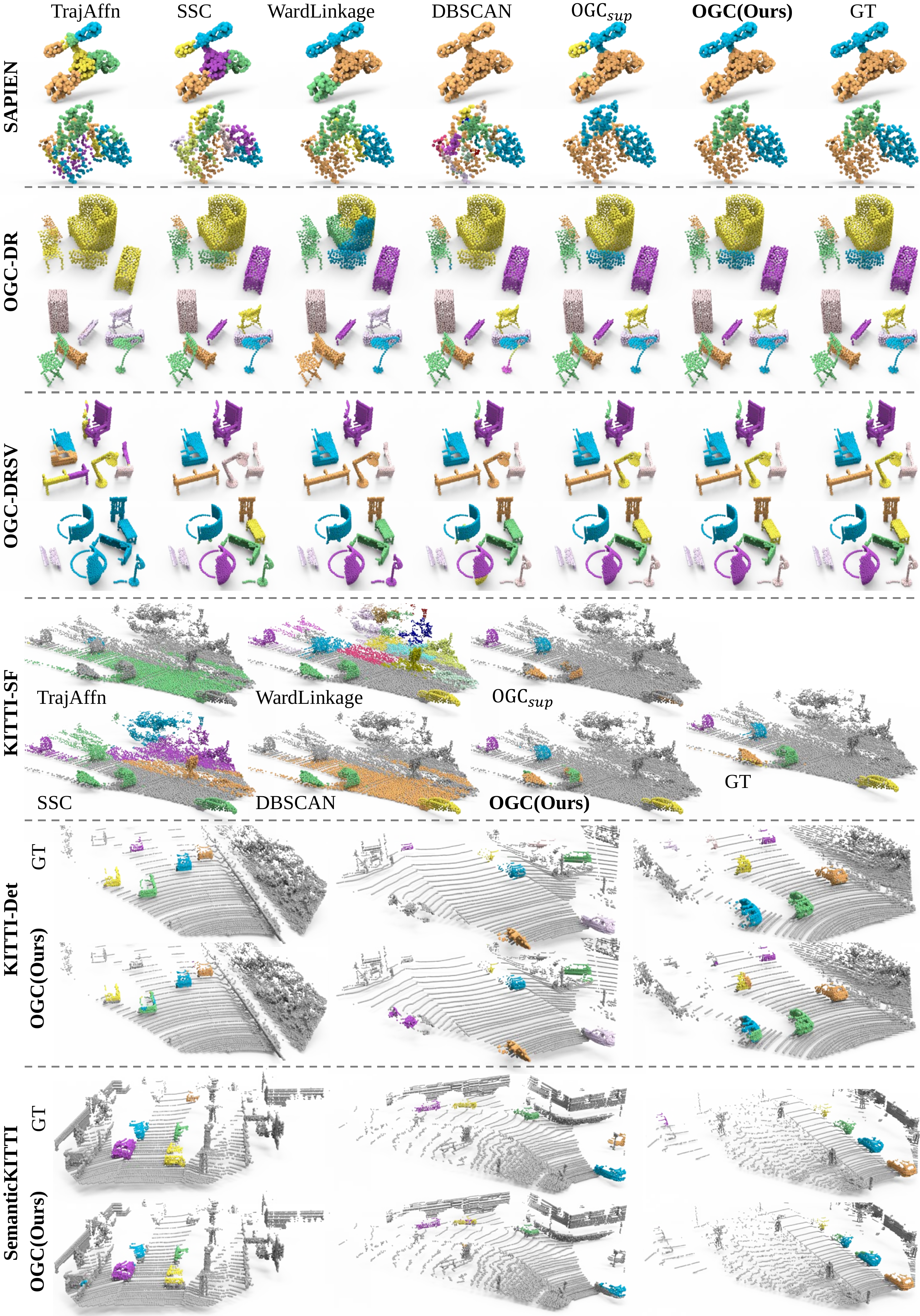}
\caption{Qualitative results on various datasets. More qualitative results can be found in Appendix \ref{sec:more_qual} and our video demo: \url{https://youtu.be/dZBjvKWJ4K0}}.
\label{fig:qual_result}
\end{figure}

In this paper, we demonstrate for the first time that 3D objects can be accurately segmented using an unsupervised method from raw point clouds. Unlike the existing approaches which usually rely on a large amount of human annotations of every 3D object for training networks, we instead turn to leverage the diverse motion patterns over sequential point clouds as supervision signals to automatically discover the objectness from single point clouds. A series of loss functions are designed to preserve the object geometry consistency over spatial and temporal scales. Extensive experiments over multiple datasets including the extremely challenging outdoor scenes demonstrate the effectiveness of our method.

\textbf{Broader Impact:} The proposed \nickname{} learns 3D objects from raw point clouds without requiring human annotations for supervision. We showcase the effectiveness for some basic applications including object part instance segmentation, indoor object segmentation and outdoor vehicle identification. We also believe that our method can be general for other domains such as AR/VR. 

\textbf{Acknowledgements:} This work was partially supported by Shenzhen Science and Technology Innovation Commission (JCYJ20210324120603011).

%% file: chaps/06_checklist.tex

\section*{Checklist}

\begin{enumerate}

\item For all authors...
\begin{enumerate}
  \item Do the main claims made in the abstract and introduction accurately reflect the paper's contributions and scope?
    \answerYes{}
  \item Did you describe the limitations of your work?
    \answerYes{See Section 5}
  \item Did you discuss any potential negative societal impacts of your work?
    \answerYes{See Section 5}
  \item Have you read the ethics review guidelines and ensured that your paper conforms to them?
    \answerYes{}
\end{enumerate}

\item If you are including theoretical results...
\begin{enumerate}
  \item Did you state the full set of assumptions of all theoretical results?
    \answerNA{}
        \item Did you include complete proofs of all theoretical results?
    \answerNA{}
\end{enumerate}

\item If you ran experiments...
\begin{enumerate}
  \item Did you include the code, data, and instructions needed to reproduce the main experimental results (either in the supplemental material or as a URL)?
    \answerYes{}
  \item Did you specify all the training details (e.g., data splits, hyperparameters, how they were chosen)?
    \answerYes{}
        \item Did you report error bars (e.g., with respect to the random seed after running experiments multiple times)?
    \answerNo{Error bars are not reported because it would be too computationally expensive}
        \item Did you include the total amount of compute and the type of resources used (e.g., type of GPUs, internal cluster, or cloud provider)?
    \answerYes{See Section 3.4}
\end{enumerate}

\item If you are using existing assets (e.g., code, data, models) or curating/releasing new assets...
\begin{enumerate}
  \item If your work uses existing assets, did you cite the creators?
    \answerYes{}
  \item Did you mention the license of the assets?
    \answerNA{}
  \item Did you include any new assets either in the supplemental material or as a URL?
    \answerNA{}
  \item Did you discuss whether and how consent was obtained from people whose data you're using/curating?
    \answerNA{}
  \item Did you discuss whether the data you are using/curating contains personally identifiable information or offensive content?
    \answerNA{}
\end{enumerate}

\item If you used crowdsourcing or conducted research with human subjects...
\begin{enumerate}
  \item Did you include the full text of instructions given to participants and screenshots, if applicable?
    \answerNA{}
  \item Did you describe any potential participant risks, with links to Institutional Review Board (IRB) approvals, if applicable?
    \answerNA{}
  \item Did you include the estimated hourly wage paid to participants and the total amount spent on participant compensation?
    \answerNA{}
\end{enumerate}

\end{enumerate}

%% file: chaps/07_app.tex
\subsection{Network Architecture}
\label{sec:net_arch}

We provide a detailed description of our object segmentation network and the auxiliary self-supervised scene flow estimator.

\textbf{(1) Object Segmentation Network}

\begin{figure}[h]
\setlength{\abovecaptionskip}{ 4 pt}
  \centering
  \includegraphics[width=1.0\linewidth]{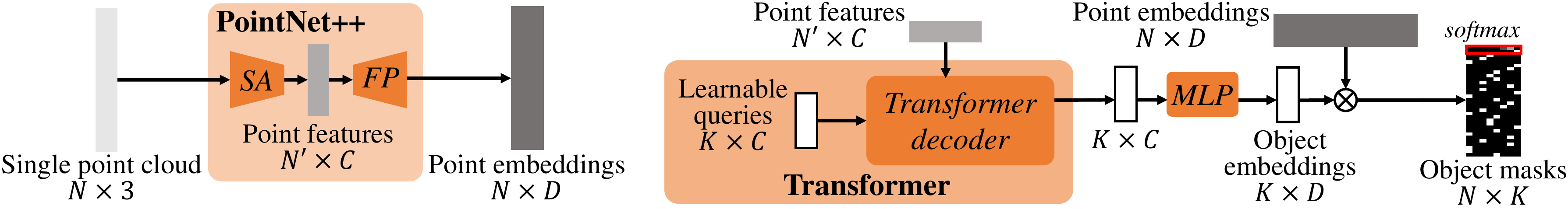}
  \caption{Detailed architecture of our object segmentation network.}
\label{fig:seg_net}
\vspace{-0.2cm}
\end{figure}

As shown in Figure \ref{fig:seg_net}, the network takes a single point cloud with $N$ points as input. It consists of Set Abstraction (SA) modules from PointNet++ \cite{Qi2017} to extract per-point features for the downsampled point cloud with $N'$ points. Feature Propagation (FP) modules are applied subsequently to obtain per-point embeddings for all $N$ points. Given the intermediate features for the $N'$ points and the $K$ learnable queries, the standard Transformer decoder \cite{Vaswani2017} is used to compute the $K$ object embeddings, each of which is expected to represent an object in the input point cloud. An MLP layer is added to reduce the dimension of object embeddings to be the same as point embeddings obtained from the PointNet++ backbone. At last, we obtain each (soft) binary mask $\boldsymbol{O}^t_k$ via a dot product between the $k^{th}$ object embedding and per-point embeddings. For each point, a softmax activation function is applied to normalize its probabilities of being assigned to different objects.

In practice, the downsampling rate and point neighborhood selection in the PointNet++ backbone are adapted to the point densities and sizes of different datasets, as shown in Table \ref{tab:config_segnet}. The embedding dimension from the Transformer decoder is set as 128 in all datasets.

\begin{table}[h]
  \tabcolsep= 0.16cm 
  \caption{Configuration of the PointNet++ backbone in our object segmentation network. $s$ denotes the point cloud downsampling/upsampling rate. $k$ controls the $K$ nearest neighbors selected within a ball with radius $r$. $c$ denotes the first input and the following output channels of MLP layers. In SA modules, the level 1-1 and 1-2 compose a multi-scale grouping (MSG) \cite{Qi2017} with outputs concatenated. In FP modules, the multi-level point features from SA are concatenated as inputs.}
  \label{tab:config_segnet}
  \centering
  \begin{tabular}{rrcccccccc}
    \toprule
    & & \multicolumn{4}{c}{SAPIEN / \nickname{}-DR / \nickname{}-DRSV} & \multicolumn{4}{c}{KITTI-SF / KITTI-Det / SemanticKITTI} \\
    & level & $s$ & $k$ & $r$ & $c$ & $s$ & $k$ & $r$ & $c$ \\
    \midrule
    \multirow{4}{*}{\makecell[c]{SA}} & 1-1 & 1/2 & 64 & 0.1(0.05) & \{3,64,64,64\} & 1/4 & 64 & 1.0 & \{3,32,32,32\} \\
    & 1-2 & 1/2 & 64 & 0.2(0.1) & \{3,64,64,128\} & 1/4 & 64 & 2.0 & \{3,32,32,64\} \\
    & 2 & 1/4 & 64 & 0.4(0.2) & \{192,128,128,256\} & 1/8 & 64 & 4.0 & \{96,64,64,128\} \\
    & 3 & & & & & 1/16 & 64 & 8.0 & \{128,128,128,256\} \\
    \hdashline
    \multirow{3}{*}{\makecell[c]{FP}} & 3 & & & & & 1/8 & & & \{384,128,128\} \\
    & 2 & 1/2 & & & \{448,256,128\} & 1/4 & & & \{224,64,64\} \\
    & 1 & 1 & & & \{131,128,128,64\} & 1 & & & \{67,64,64,64\} \\
    \bottomrule
  \end{tabular}
\end{table}

\textbf{(2) Self-Supervised Scene Flow Estimator}

We use the existing FlowStep3D as our self-supervised scene flow estimator. This method extracts per-point features via a PointNet++ backbone from two input point cloud frames separately. Then it adopts a recurrent architecture to refine the scene flow predictions iteratively. We refer readers to \cite{Kittenplon2021} for more details. 
On SAPIEN and \nickname{}-DR / \nickname{}-DRSV datasets, with smaller scene sizes and fewer input points, we remove the last SA module with 1/32 downsampling rate and reduce the number of nearest neighbors as half of its original choice in all modules.

\clearpage
\subsection{Object-Aware ICP Algorithm}
\label{sec:oa_icp}

In Algorithm \ref{alg:oa_icp}, we present our object-aware ICP (Iterative Closest Point) algorithm.

\vspace{-0.4cm}
\begin{algorithm}[H]
\caption{Object-aware ICP algorithm. 
Assume each training sample contains a pair of point clouds and scene flow estimations $(\boldsymbol{P}^t, \boldsymbol{P}^{t+1}, \boldsymbol{M}^t \in \mathbb{R}^{N\times 3})$, and object masks $(\boldsymbol{O}^t, \boldsymbol{O}^{t+1} \in \mathbb{R}^{N\times K})$ obtained from a trained object segmentation network.}
\label{alg:oa_icp}
\begin{algorithmic}

\STATE{\textit{Step 1: Match the individual masks in $\boldsymbol{O}^t$ and $\boldsymbol{O}^{t+1}$.}}
\STATE{\phantom{xx}$\bullet$ Use the estimated scene flows to warp the first point cloud: $\boldsymbol{P}^t_{w} = \boldsymbol{P}^t + \boldsymbol{M}^t$, the warped $\boldsymbol{P}^t_{w}$ naturally inherits the per-point object masks from $\boldsymbol{P}^t$: $\boldsymbol{O}^t_{w} = \boldsymbol{O}^t$.}
\STATE{\phantom{xx}$\bullet$ Compute another object masks $\boldsymbol{\hat{O}}^{t+1}$ for the second point cloud $\boldsymbol{P}^{t+1}$ using the nearest-neighbor interpolation from $(\boldsymbol{P}^{t}_{w}, \boldsymbol{O}^t_{w})$.}
\STATE{\phantom{xx}$\bullet$ Leverage the Hungarian algorithm \cite{Kuhn1955} to one-one match the individual masks in $\boldsymbol{\hat{O}}^{t+1}$ and $\boldsymbol{O}^{t+1}$ according to the object pair-wise IoU scores.}
\STATE{\phantom{xx}$\bullet$ Reorder the masks in $\boldsymbol{O}^{t+1}$ to align with $\boldsymbol{\hat{O}}^{t+1}$, thus aligh with $\boldsymbol{O}^t$.}\\

\STATE{\textit{Step 2: Iteratively refine the rigid scene flow estimations}}
\STATE{\phantom{xx}$\bullet$ Compute the per-point object consistency scores $\boldsymbol{O} \in \mathbb{R}^{N \times N}$ between $\boldsymbol{P}^t$ and $\boldsymbol{P}^{t+1}$: $\boldsymbol{O} = \boldsymbol{O}^t (\boldsymbol{O}^{t+1})^{\intercal}$.}\\

\FOR {number of iterations $I$}{} 

\STATE{\phantom{xx}$\bullet$ Compute the per-point soft correspondence scores $\boldsymbol{C} \in \mathbb{R}^{N \times N}$ between $\boldsymbol{P}^t$ and $\boldsymbol{P}^{t+1}$ based on the nearest-neighbor (closest point) distances, where 
\begin{equation*}
\boldsymbol{C}_{ij} = exp(- \boldsymbol{\delta}_{ij} / \tau), \boldsymbol{\delta}_{ij} = \Big\| \boldsymbol{p}^t_i + \boldsymbol{m}^t_i - \boldsymbol{p}^{t+1}_j \Big\|_2
\end{equation*}
}

\STATE{\phantom{xx}$\bullet$ Filter the per-point correspondence scores by the object consistency scores: $\boldsymbol{C} = \boldsymbol{C} * \boldsymbol{O}$.}
\STATE{\phantom{xx}$\bullet$ Update the scene flows $\boldsymbol{M}^t$ from the object-aware soft correspondences, where 
\begin{equation*}
\boldsymbol{m}^t_i = \dfrac{ \sum_{j=1}^{N} \boldsymbol{C}_{ij} (\boldsymbol{p}^{t+1}_j - \boldsymbol{p}^{t}_i) }{ \sum_{j=1}^{N} \boldsymbol{C}_{ij} }
\end{equation*}
}

\STATE{\phantom{xx}$\bullet$ For the $k^{th}$ object, retrieve its (soft) binary mask $\boldsymbol{O}^t_k$, and then feed the tuple $\{\boldsymbol{P}^t, \boldsymbol{P}^{t}+\boldsymbol{M}^t, \boldsymbol{O}^t_k\}$ into weighted-Kabsch \cite{kabsch,Gojcic2020} algorithm, estimating its transformation matrix $\boldsymbol{T}_k$.}
\STATE{\phantom{xx}$\bullet$ Update the scene flows $\boldsymbol{M}^t$ from the estimated transformations: 
\begin{equation*}
\boldsymbol{M}^t = \sum_{k=1}^K \boldsymbol{O}_k^t * ( \boldsymbol{T}_k \circ \boldsymbol{P}^t - \boldsymbol{P}^t )
\end{equation*}
}

\ENDFOR
\\
Return the scene flows $\boldsymbol{M}^t$ from the last iteration.
\end{algorithmic}
\end{algorithm}
\vspace{-0.8cm}

In the iterative optimization, as the scene flow estimations gradually approach more consistent and accurate values, the number of iterations $I$ in the object-aware ICP can be reduced for efficiency. We set $I$ in the object-aware ICP as \{20, 10, 5\} in the round \{1, 2, 3\} of the iterative optimization.

\setlength{\abovecaptionskip}{ -10 pt}
\begin{wraptable}{r}{9cm}
  \tabcolsep= 0.08cm 
  \caption{Scene flow estimation on KITTI-SF benchmark.}
  \label{tab:compare_kabsch}
  \centering
  \begin{tabular}{rllll}
    \toprule
    & EPE3D$\downarrow$ & AccS$\uparrow$ & AccR$\uparrow$ & Outlier$\downarrow$ \\
    \midrule
    FlowStep3D \cite{Kittenplon2021} & 10.21 & 70.80 & 83.94 & 24.56 \\
    Weighted Kabsch \cite{kabsch,Gojcic2020} & 9.31 & 71.01 & 81.20 & 28.75 \\
    \textbf{Object-aware ICP} & \textbf{6.72} & \textbf{80.16} & \textbf{89.08} & \textbf{22.56} \\
    \bottomrule
  \end{tabular}
\vspace{-0.3 cm}
\end{wraptable}

Compared to the weighted-Kabsch \cite{kabsch,Gojcic2020} algorithm, 
our object-aware ICP algorithm takes two frames as input to correct the inconsistency in the flows. As shown in Table \ref{tab:compare_kabsch}, our algorithm obtains a larger improvement for scene flow estimations. In general, our algorithm is an extension of the classical ICP \cite{Arun1987} to 3D scenes with multiple rigid objects. Our algorithm can be naturally implemented in a batch-wise manner, without sacrificing the optimization speed of the network.

\subsection{\nickname{}-DR and \nickname{}-DRSV Datasets}
\label{sec:ogcdr_datagen}

Here we provide details about the data generation of our \nickname{}-DR and \nickname{}-DRSV datasets. Following \cite{Peng2020a}, we first generate 5000 static scenes with 4 $\sim$ 8 objects. For a single scene, the ratio of width and length of the ground plane is uniformly sampled between $0.6\sim 1.0$. For each object in a scene, its scale is sampled from $0.2\sim 0.45$. The object rotation angles around the vertical y-axis are randomly sampled from $-180^{\circ}\sim 180^{\circ}$. Unlike \cite{Peng2020a}, we do not keep walls and ground planes in the generated raw point clouds since these textureless surfaces create intractable ambiguities for self-supervised scene flow estimation. In fact, it is trivial to detect and remove them via plane fitting in real-world indoor scenes. The walls and ground planes are simply used to place all objects in a more realistic manner.

We then create rigid dynamics for the objects. For each object in a scene, we sample a rigid transformation relative to its pose in the previous frame. In particular, we first uniformly sample an angle from $-10^{\circ}\sim 10^{\circ}$ rotated by y/x/z axis with the probability of \{0.6, 0.2, 0.2\} respectively. Afterwards, we uniformly sample a translation only on the x-z plane from the range of $-0.04\sim 0.04$ for each object, ensuring that the object is always on the ground. Note that, we reject all samples which have objects overlapped or being out of the scene boundary.

\begin{figure}[t]
\setlength{\abovecaptionskip}{ 4 pt}
  \centering
  \includegraphics[width=1.0\linewidth]{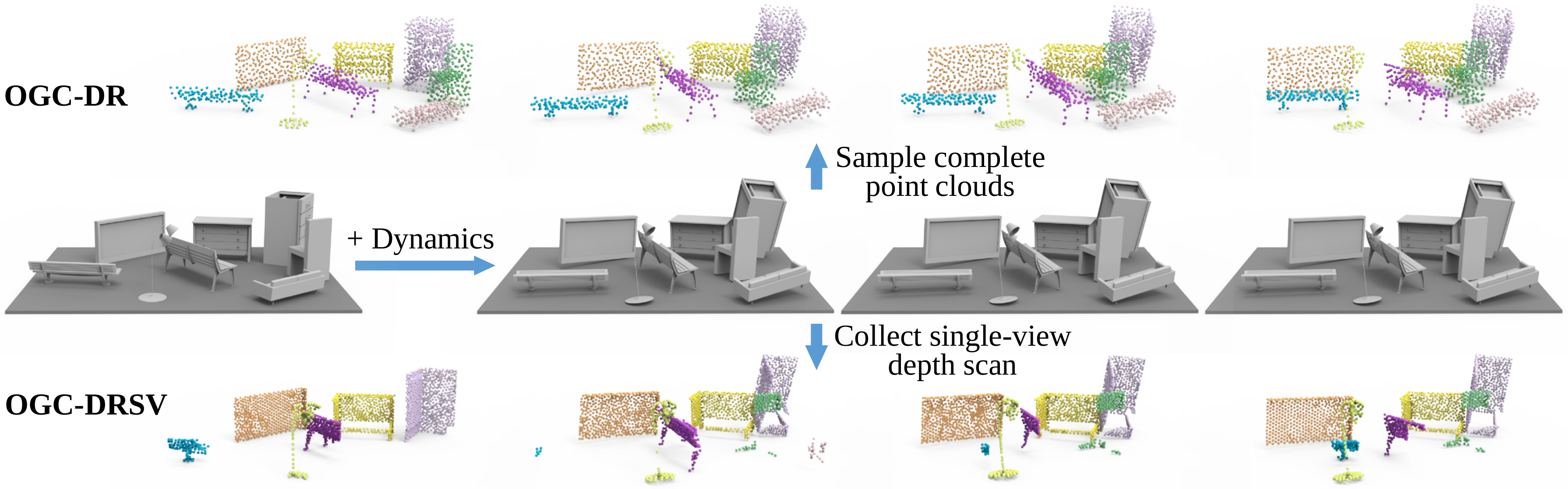}
  \caption{Illustration of the data generation process for our \nickname{}-DR dataset.}
\label{fig:dataset_gen}
\end{figure}

The last point cloud sampling step varies for the two datasets. For \nickname{}-DR, we directly sample point clouds from the surfaces of complete mesh models, while for \nickname{}-DRSV, we collect single-view depth scans on mesh models. Note that for both datasets, the point sampling is independently conducted on each frame in a scene. Therefore, there is no exact point correspondences between consecutive frames. This setting is consistent with the scene flow estimation task in general real-world scenes. Figure \ref{fig:dataset_gen} illustrates the complete data generation process.

\subsection{Additional Implementation Details}
\label{sec:more_implement}

\phantom{xxx}\textbf{(1) Data Preparation}

\textbf{SAPIEN:} In SAPIEN, each scene (an articulated object) has 4 sequential scans. During  training, we leverage consecutive frame pairs (both forward and backward) only, because the self-supervised scene flow estimator can hardly handle rapid motions. Therefore, each object has 6 pairs of point clouds. Given 13682/2356 objects in training/validation splits, we get 82092 training and 14136 validation frame pairs. The 720 objects in testing split contribute 2880 individual frames for evaluation.

\textbf{\nickname{}-DR/\nickname{}-DRSV:} Similar to SAPIEN, each scene in \nickname{}-DR/\nickname{}-DRSV holds 4 sequential frames. The 3750/250 scenes in training/validation splits give 22500 training and 1500 validation frame pairs, and the 1000 scenes in the testing split provide 4000 testing frames.

\textbf{KITTI-SF:} The 100 pairs of point clouds in the training split of KITTI-SF contribute 200 training pairs (both forward and backward), and the other 100 pairs in testing split provide 200 individual frames for evaluation. A tricky problem in KTTI-SF is the scene flow estimation for the ground. The textureless ground poses intractable ambiguities for the self-supervised scene flow estimator. However, we cannot simply remove the ground by applying a height threshold, because the background points above the ground will no longer be spatially connected, thus breaking the assumption behind our geometry smoothness regularizer $\ell_{smooth}$. To address this issue, we apply the self-supervised scene flow estimator to points above the ground only. Meanwhile, we use the classical ICP \cite{Arun1987} algorithm onto points above the ground and regard the fitted transformation as motions for ground points (\ie, the camera ego-motion). Although this solution relies on an assumption that static background points dominate the scene, our object-aware ICP algorithm can empirically alleviate potential errors in the iterative optimization.

\phantom{xxx}\textbf{(2) Hyperparameter Selection}

\textbf{Geometry Smoothness Regularization:} We choose $L1$ for the distance function $d()$ given its less sensitivity to "outliers", \ie, adjacent points belonging to different objects. 
As shown in Table \ref{tab:hparam_smooth}, we select two groups of neighboring points from two different scales (denoted by \{$k_1$, $r_1$\} and \{$k_2$, $r_2$\}) and weight them by \{3.0, 1.0\} in $\ell_{smooth}$.

\textbf{Geometry Invariance Loss:} For the distance function $\hat{d}()$ here, $L1$, $L2$ and cross-entropy are all theoretically reasonable. We choose $L2$ for the best performance. The transformation for augmentation comprises a scale factor uniformly sampled from $0.95\sim 1.05$ and a rotation around the vertical y-axis sampled from $-180\sim 180^\circ$. On KITTI-SF dataset, we add an x-z translation sampled from $-1\sim 1$ and a y translation sampled from $-0.1\sim 0.1$.

\setlength{\abovecaptionskip}{ -10 pt}
\setlength{\belowdisplayskip}{8pt}
\begin{wraptable}{r}{7cm}
  \tabcolsep= 0.13cm 
  \caption{The choices of neighboring points in the geometry smoothness regularization. $k$ controls the $K$ nearest neighbors selected within a ball with radius $r$.}
  \label{tab:hparam_smooth}
  \centering
  \begin{tabular}{cccc|cccc}
    \hline
    \multicolumn{4}{c}{SAPIEN / \nickname{}-DR} & \multicolumn{4}{c}{KITTI-SF} \\
    $k_1$ & $r_1$ & $k_2$ & $r_2$ & $k_1$ & $r_1$ & $k_2$ & $r_2$ \\
    \hline
    8 & 0.1(0.02) & 16 & 0.2(0.04) & 32 & 1.0 & 64 & 2.0 \\
    \hline
  \end{tabular}
\vspace{-0.3 cm}
\end{wraptable}

\textbf{Network Training:} We adopt the Adam optimizer with a learning rate of 0.001 and train on SAPIEN/\nickname{}-DR/KITTI-SF datasets for 40/40/200 epochs, respectively. The batch size is set as 32/8/4 on each dataset to fill in the whole memory of a single RTX3090 GPU. The three losses $\ell_{dynamic}$, $\ell_{smooth}$ and $\ell_{invariant}$ are weighted by \{10.0, 0.1, 0.1\}. On SAPIEN and \nickname{}-DR, since $\ell_{invariant}$ can slow down the convergence, we first train 20 epochs without it and then add it back. We also find that $\ell_{smooth}$ occasionally overwhelms the initial iterations of training, causing network predictions to collapse and assign all points to a single object. Therefore, we empirically disable $\ell_{smooth}$ before iterating through the initial 2000/2000/200 samples on SAPIEN/\nickname{}-DR/KITTI-SF datasets.

\phantom{xxx}\textbf{(3) Baseline methods on KITTI-SF} 

\begin{figure}[h]
\setlength{\abovecaptionskip}{ 4 pt}
  \centering
  \includegraphics[width=1.0\linewidth]{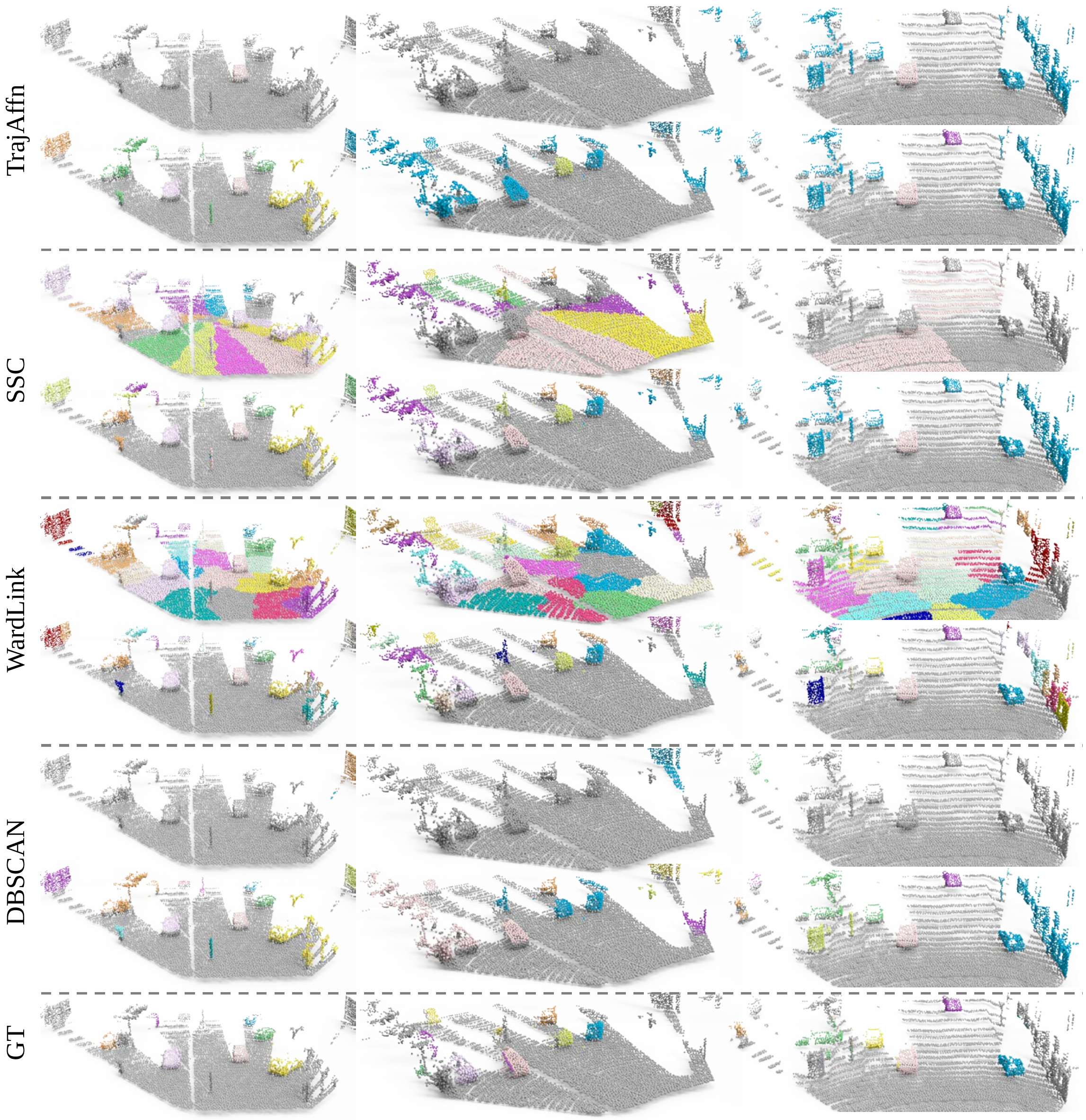}
  \caption{Qualitative results for ablation studies about the ground plane prior on KITTI-SF. For each baseline method, segmentation results without (top row) and with (bottom row) the ground plane prior are shown.}
\label{fig:kittisf_baselines}
\vspace{-0.3cm}
\end{figure}

Since the KITTI-SF dataset is too challeging for the classical unsupervised methods, we leverage the prior about ground planes in the dataset to improve baseline methods. First, we detect and temporarily remove the ground plane, letting the baseline algorithms to segment above-ground points only. For TrajAffn and SSC, we can use motion information to merge the ground points with above-ground segments that are likely to be part of the static background. To do this, we employ the Kabsch algorithm to estimate the rigid transformation of the ground. Then the above-ground segments whose motions are well fitted by the ground's transformation will be incorporated. For WardLinkage and DBSCAN, the ground is treated as a separate segment. We conduct ablation studies to validate the use of ground plane prior for these baseline methods on the KITTI-SF dataset, as shown in Table \ref{tab:res_kittisf_groundprior} and Figure \ref{fig:kittisf_baselines}. After using the ground plane prior, SSC and WardLinkage gain remarkable improvements both quantitatively and qualitatively. For TrajAffn and DBSCAN, although the quantitative performance gain is not significant, we find that their qualitative results become more meaningful.

\begin{table}[h]
      \tabcolsep= 0.18cm 
  \caption{Ablation studies about the ground plane prior for baseline methods on KITTI-SF.}
  \label{tab:res_kittisf_groundprior}
  \centering
  \begin{tabular}{rcccccccc}
    \toprule
    & use ground plane prior & AP$\uparrow$ & PQ$\uparrow$ & F1$\uparrow$ & Pre$\uparrow$ & Rec$\uparrow$ & mIoU$\uparrow$ & RI$\uparrow$ \\
    \midrule
    \multirow{2}{*}{\makecell[c]{TrajAffn \cite{Ochs2014}}} & & 30.4 & 34.7 & 42.7 & 40.5 & 45.3 & 49.0 & 83.5 \\
    & \checkmark & 24.0 & 30.2 & 43.2 & 37.6 & 50.8 & 48.1 & 58.5 \\
    \hdashline
    \multirow{2}{*}{\makecell[c]{SSC \cite{Nunes2018}}} & & 2.9 & 5.2 & 7.5 & 6.2 & 9.5 & 19.3 & 33.2 \\
    & \checkmark & 12.5 & 20.4 & 28.4 & 22.8 & 37.6 & 41.5 & 48.9 \\
    \hdashline
    \multirow{2}{*}{\makecell[c]{WardLinkage \cite{Jr.1963}}} & & 1.3 & 2.4 & 3.8 & 2.2 & 14.3 & 26.8 & 15.7 \\
    & \checkmark & 25.0 & 16.3 & 22.9 & 13.7 & 69.8 & 60.5 & 44.9 \\
    \hdashline
    \multirow{2}{*}{\makecell[c]{DBSCAN \cite{Ester1996}}} & & 14.8 & 29.9 & 32.9 & 46.5 & 25.4 & 31.3 & 84.8 \\
    & \checkmark & 13.4 & 22.8 & 32.6 & 26.7 & 42.0 & 42.6 & 55.3 \\
    \bottomrule
  \end{tabular}
    \vspace{-0.4cm}
\end{table}

\subsection{Additional Ablation Studies}
\label{sec:more_abla}

\phantom{xxx}\textbf{(1) Geometry Consistency Losses}

We conduct additional ablation experiments on the curated SAPIEN dataset for a more comprehensive analysis of losses in our framework. Recall that without the dynamic rigid loss $\ell_{dynamic}$, network predictions collapse and assign all points to a single object. The full SAPIEN dataset holds a number of point cloud frames with only 2 or 3 object parts, enabling the ablated model without $\ell_{dynamic}$ to still get plausible scores. However, as shown in Tables \ref{tab:abl_sapien2}\&\ref{tab:abl_sapien3}, once we evaluate the ablated models on point clouds with $\geq 3$ object parts (Table \ref{tab:abl_sapien2}) or $\geq 4$ object parts (Table \ref{tab:abl_sapien3}), the performance of the ablated model without $\ell_{dynamic}$ drops rapidly. This clearly shows that the $\ell_{dynamic}$ loss is truly critical to tackle complex scenes with more and more objects. Figure \ref{fig:qual_abl} shows qualitative examples.

\setlength{\abovecaptionskip}{ -10 pt}
\begin{table}[h]
  \tabcolsep= 0.2cm 
  \caption{Additional ablation results on a curated SAPIEN dataset where only point clouds with $\geq 3$ object parts are kept (844 frames in total).}
  \label{tab:abl_sapien2}
  \centering
  \begin{tabular}{rccccccc}
    \toprule
     Config. & AP$\uparrow$ & PQ$\uparrow$ & F1$\uparrow$ & Pre$\uparrow$ & Rec$\uparrow$ & mIoU$\uparrow$ & RI$\uparrow$ \\
    \midrule
    w/o $\ell_{dynamic}$ & 15.8 & 21.0 & 32.7 & \textbf{69.7} & 21.4 & 18.4 & 44.1 \\
    w/o $\ell_{smooth}$ & 13.0 & 15.5 & 23.7 & 18.9 & 31.8 & 42.9 & 65.5 \\
    w/o $\ell_{invariant}$ & 23.8 & 28.3 & 41.3 & 47.6 & 36.5 & 38.2 & 64.0 \\
    Full \nickname{} & \textbf{30.8} & \textbf{34.0} & \textbf{48.2} & 52.4 & \textbf{44.6} & \textbf{43.4} & \textbf{67.4} \\
    \bottomrule
  \end{tabular}
\end{table}
\setlength{\abovecaptionskip}{ -10 pt}
\begin{table}[ht]
  \tabcolsep= 0.2cm 
  \caption{Additional ablation results on a curated SAPIEN dataset where only point clouds with $\geq 4$ object parts are kept (120 frames in total).}
  \label{tab:abl_sapien3}
  \centering
  \begin{tabular}{rccccccc}
    \toprule
     Config. & AP$\uparrow$ & PQ$\uparrow$ & F1$\uparrow$ & Pre$\uparrow$ & Rec$\uparrow$ & mIoU$\uparrow$ & RI$\uparrow$ \\
    \midrule
    w/o $\ell_{dynamic}$ & 10.8 & 12.9 & 22.4 & \textbf{65.0} & 13.5 & 11.1 & 35.0 \\
    w/o $\ell_{smooth}$ & 12.8 & 13.9 & 22.2 & 20.3 & 24.5 & \textbf{35.3} & \textbf{67.3} \\
    w/o $\ell_{invariant}$ & 15.7 & 21.6 & 31.9 & 46.2 & 24.3 & 26.8 & 60.3 \\
    Full \nickname{} & \textbf{22.3} & \textbf{26.6} & \textbf{40.1} & 55.0 & \textbf{31.6} & 29.7 & 59.8 \\
    \bottomrule
  \end{tabular}
\end{table}
\setlength{\abovecaptionskip}{ -10 pt}
\begin{table}[ht]
  \caption{Additional ablation studies about loss designs on KITTI-SF.}
  \label{tab:abl_kitti}
  \centering
  \begin{tabular}{rccccccc}
    \toprule
     Config. & AP$\uparrow$ & PQ$\uparrow$ & F1$\uparrow$ & Pre$\uparrow$ & Rec$\uparrow$ & mIoU$\uparrow$ & RI$\uparrow$ \\
    \midrule
    w/o $\ell_{dynamic}$ & 24.8 & 37.1 & 39.6 & \textbf{100.0} & 24.7 & 31.3 & 88.3 \\
    w/o $\ell_{smooth}$ & 44.9 & 31.8 & 39.6 & 30.9 & 55.1 & 61.2 & 90.2 \\
    w/o $\ell_{invariant}$ & 47.1 & 35.0 & 43.0 & 35.5 & 54.4 & 60.7 & 92.5 \\
    Full \nickname{} & \textbf{54.4} & \textbf{42.4} & \textbf{52.4} & 47.3 & \textbf{58.8} & \textbf{63.7} & \textbf{93.6} \\
    \bottomrule
  \end{tabular}
\end{table}

\begin{figure}[t]
\setlength{\abovecaptionskip}{ 4 pt}
  \centering
  \includegraphics[width=1.0\linewidth]{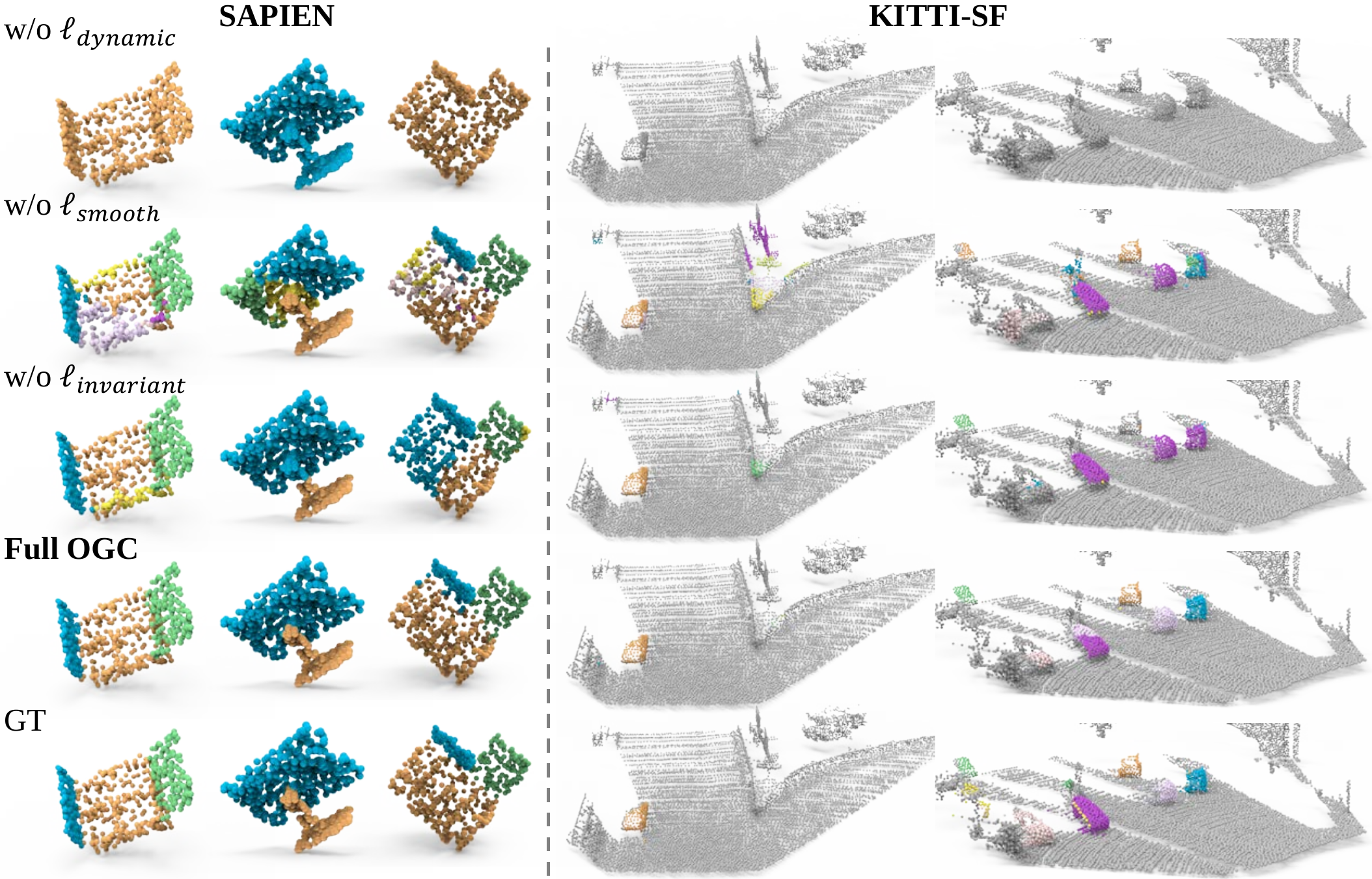}
  \caption{Qualitative results for ablation study on SAPIEN and KITTI-SF.}
\label{fig:qual_abl}
\end{figure}

We also conduct additional ablative experiments on the KITTI-SF dataset, as shown in Table \ref{tab:abl_kitti}. Similar to the results on the SAPIEN dataset, we observe the collapse of object segmentation without $\ell_{dynamic}$ and the oversegmentation issue without $\ell_{smooth}$. Figure \ref{fig:qual_abl} gives an intuitive illustration.

\clearpage
\phantom{xxx}\textbf{(2) Use of the Invariance Loss in Iterative Optimization}

We conduct ablation experiments on the KITTI-SF dataset to validate our choice of using the invariance loss $l_{invariant}$ only in the final round of iterative optimization. To do this, we iteratively optimize on KITTI-SF for 2 rounds, with two different configurations: (i) We use $l_{invariant}$ in object segmentation optimization of the two rounds. (ii) We use $l_{invariant}$ only in the 2nd round. For each configuration, we run for five times with different random seeds and report the mean results with uncertainty levels.

\begin{table}[h]
      \tabcolsep= 0.2cm 
  \caption{Ablation results about the use of the invariance loss in iterative optimization. \#$R$ denotes the number of iterative optimization rounds.}
  \label{tab:invloss_iter}
  \centering
  \scriptsize
  \begin{tabular}{ccc|ccccccc} 
    \hline
    \#$R$ & Split & Config & AP$\uparrow$ & PQ$\uparrow$ & F1$\uparrow$ & Pre$\uparrow$ & Rec$\uparrow$ & mIoU$\uparrow$ & RI$\uparrow$ \\ 
    \hline
    \multirow{4}{*}{\makecell[c]{1}} & \multirow{2}{*}{\makecell[c]{Train}} & (i) & 40.7$\pm$2.1 & 27.7$\pm$0.8 & 38.7$\pm$1.1 & 27.0$\pm$0.9 & 68.0$\pm$1.6 & 61.5$\pm$0.8 & 62.9$\pm$1.1 \\
    & & (ii) & 41.4$\pm$3.0 & 31.3$\pm$0.9 & 43.5$\pm$1.2 & 30.5$\pm$1.1 & 75.4$\pm$1.2 & 64.8$\pm$0.7 & 60.6$\pm$1.0 \\
    \cline{2-10}
    & \multirow{2}{*}{\makecell[c]{Test}} & (i) & 34.1$\pm$2.0 & 24.1$\pm$1.6 & 35.0$\pm$2.1 & 26.0$\pm$1.8 & 53.9$\pm$2.2 & 52.9$\pm$1.1 & 57.2$\pm$0.9 \\
    & & (ii) & 29.1$\pm$1.9 & 22.7$\pm$1.0 & 33.7$\pm$1.6 & 24.9$\pm$1.4 & 52.1$\pm$1.6 & 51.2$\pm$0.8 & 54.2$\pm$1.1 \\
    \hline
    \multirow{4}{*}{\makecell[c]{2}} & \multirow{2}{*}{\makecell[c]{Train}} & (i) & 57.0$\pm$8.1 & 41.3$\pm$7.3 & 52.4$\pm$7.2 & 41.9$\pm$9.2 & 71.7$\pm$2.1 & 69.2$\pm$3.6 & 83.1$\pm$11.1 \\
    & & (ii) & 67.0$\pm$2.3 & 50.5$\pm$2.3 & 61.5$\pm$2.4 & 53.1$\pm$3.6 & 73.1$\pm$1.4 & 73.7$\pm$0.9 & 95.5$\pm$0.3 \\
    \cline{2-10}
    & \multirow{2}{*}{\makecell[c]{Test}} & (i) & 41.9$\pm$8.2 & 33.0$\pm$6.4 & 43.0$\pm$6.1 & 35.8$\pm$7.6 & 54.7$\pm$3.0 & 58.4$\pm$4.0 & 79.4$\pm$13.0 \\
    & & (ii) & 51.8$\pm$2.2 & 40.8$\pm$2.2 & 50.6$\pm$2.6 & 45.5$\pm$3.5 & 57.2$\pm$1.7 & 62.1$\pm$1.2 & 93.4$\pm$0.3 \\
    \hline
  \end{tabular}
  \vspace{-0.3 cm}
\end{table}

\textbf{Analysis:} As shown in Table \ref{tab:invloss_iter}, in the 1st round, the model (ii) trained without $l_{invariant}$ sacrifices some generalization performance on testing data (F1: 33.7 $vs$ 35.0) while fitting better on training data (F1: 43.5 $vs$ 38.7). As shown in Table \ref{tab:invloss_iter_flow}, such advantages in segmentation performance on training data lead to more refined scene flows (EPE3D: 2.36 $vs$ 3.12), which will directly influence the optimization in the next round. In constrast, better segmentation on testing data cannot be passed to the next round. In the 2nd round, both models are trained with $l_{invariant}$. The model (ii) stably produces superior segmentation results on both training (F1: 61.5$\pm$2.4 $vs$ 52.4$\pm$7.2) and testing data (F1: 50.6$\pm$2.6 $vs$ 43.0$\pm$6.1), owing to higher quality scene flow refinement inherited from the previous round.

\setlength{\abovecaptionskip}{ -10 pt}
\begin{wraptable}{r}{7.5cm}
  \tabcolsep= 0.08cm 
  \caption{Refined scene flow estimation after the 1st round on KITTI-SF dataset.}
  \label{tab:invloss_iter_flow}
  \centering
  \begin{tabular}{ccccc}
    \toprule
    Config & EPE3D$\downarrow$ & AccS$\uparrow$ & AccR$\uparrow$ & Outlier$\downarrow$ \\
    \midrule
    (i) & 3.12$\pm$0.16 & 92.8$\pm$0.7 & 94.7$\pm$0.5 & 23.9$\pm$0.5 \\
    (ii) & 2.36$\pm$0.12 & 94.7$\pm$0.5 & 96.4$\pm$0.2 & 22.5$\pm$0.1 \\
    \bottomrule
  \end{tabular}
\vspace{-0.3 cm}
\end{wraptable}

The findings above are consistent with our theoretical expectations. In the iterative optimization, the previous rounds can only influence the final results by passing refined scene flows (on training split) to the following rounds. The invariance loss $l_{invariant}$ brings better generalization, especially to static objects, while these properties are of little use for scene flow refinement on training data. Therefore, in the previous rounds of iterative optimization, we can exclude $l_{invariant}$ and let the model focus on moving objects in training samples and produce more refined scene flows. In the last round, $l_{invariant}$ can be included to boost the generalization ability of the final model.

\phantom{xxx}\textbf{(3) Robustness to Scene Flow Distortions}

We investigate the robustness of our method to scene flow distortions on the OGC-DR and KITTI-SF datasets. To do this, we conduct experiments on three types of distorted scene flows: 
(i) We add different degrees of zero-mean Gaussian noise into the ground truth scene flows, and these noisy scene flows are used to supervised our object segmentation network.
(ii) We use insufficiently trained scene flow estimators to produce low-quality scene flow estimations for supervision.
(iii) We use the initial scene flow estimations which has not been refined by our iterative optimization. On our OGC-DR dataset, we evaluate all these ablations. On the KITTI-SF dataset, we only evaluate (i) and (iii), as we use a FlowStep3D model publicly released by the authors to estimate scene flows for KITTI-SF.
The intermediate training models are not available to evaluate (ii).

\begin{table}[h]
      \tabcolsep= 0.19cm 
  \caption{Ablation results about the robustness to scene flow distortions on OGC-DR. Bold text denotes \textbf{the configuration of full OGC.} \#$R$ denotes the number of iterative optimization rounds. We report the object segmentation performance on the testing set and scene flow quality on training set (the scene flow quality of the testing set is irrelevant to our object segmentation).}
  \label{tab:robust_ogcdr}
  \centering
  \scriptsize
  \begin{tabular}{ccl|ccccccc|cc} 
    \hline
    & & & \multicolumn{7}{c}{Object Segmentation} & \multicolumn{2}{c}{Scene Flow} \\
    & Flow Source & \#$R$ & AP$\uparrow$ & PQ$\uparrow$ & F1$\uparrow$ & Pre$\uparrow$ & Rec$\uparrow$ & mIoU$\uparrow$ & RI$\uparrow$ & EPE3D$\downarrow$ & AccR$\uparrow$ \\ 
    \hline
    \multirow{3}{*}{\makecell[c]{Ablation (i)}} & GT + Gaussian (std=1.0) & 1 & 91.3 & 83.3 & 88.1 & 84.4 & 92.1 & 89.2 & 97.6 & 1.60 & 73.9 \\
    & GT + Gaussian (std=2.0) & 1 & 86.4 & 79.7 & 86.2 & 85.0 & 87.4 & 83.8 & 96.5 & 3.19 & 19.9 \\
    & GT + Gaussian (std=3.0) & 1 & 85.2 & 77.2 & 84.5 & 82.9 & 86.2 & 82.0 & 95.8 & 4.79 & 6.9 \\
    \hline
    \multirow{3}{*}{\makecell[c]{Ablation (ii)}} & FlowStep3D (epoch=20) & 1 & 90.1 & 81.0 & 86.4 & 81.6 & 91.9 & 88.2 & 96.8 & 1.14 & 90.2 \\
    & FlowStep3D (epoch=10) & 1 & 89.1 & 79.8 & 86.1 & 82.1 & 90.6 & 86.2 & 96.3 & 1.45 & 86.0 \\
    & FlowStep3D (epoch=1) & 1 & 84.7 & 71.6 & 80.5 & 74.3 & 87.8 & 80.9 & 93.5 & 3.84 & 32.2 \\
    \hline
    \multirow{2}{*}{\makecell[c]{Ablation (iii)}} & FlowStep3D (epoch=50) & 1 & 91.3 & 83.7 & 88.4 & 84.5 & 92.7 & 89.7 & 97.6 & 0.98 & 90.0 \\
    & \textbf{FlowStep3D (epoch=50)} & \textbf{2} & 92.3 & 85.1 & 89.4 & 85.6 & 93.6 & 90.8 & 97.8 & 0.76 & 92.2 \\
    \hline
  \end{tabular}
  \vspace{-0.3cm}
\end{table}

\paragraph{Analysis on OGC-DR:} As shown in Table \ref{tab:robust_ogcdr}, our OGC is robust to Gaussian noises in scene flows. The model maintains 85.2 AP even when the AccR of scene flows degrades to 6.9 only. In contrast, the flow distortions from insufficiently trained estimators incur a notable drop in the segmentation performance. The AP drops to 84.7 even when the scene flow AccR is still 32.2. From this, we hypothesize that our OGC can be robust to noisy flows with large variance thanks to the rigid loss integrated with weighted-Kabsch algorithm, but sensitive to large biases in estimated scene flows.

\begin{table}[h]
      \tabcolsep= 0.19cm 
  \caption{Ablation results about the robustness to scene flow distortions on KITTI-SF. Bold text denotes \textbf{the configuration of full OGC.}}
  \label{tab:robust_kittisf}
  \centering
  \scriptsize
  \begin{tabular}{ccl|ccccccc|cc}
    \hline
    & & & \multicolumn{7}{c}{Object Segmentation} & \multicolumn{2}{c}{Scene Flow} \\
    & Flow Source & \#$R$ & AP$\uparrow$ & PQ$\uparrow$ & F1$\uparrow$ & Pre$\uparrow$ & Rec$\uparrow$ & mIoU$\uparrow$ & RI$\uparrow$ & EPE3D$\downarrow$ & AccR$\uparrow$ \\ 
    \hline
    \multirow{2}{*}{\makecell[c]{Ablation (i)}} & GT + Gaussian (std=10.0) & 1 & 61.1 & 49.5 & 59.5 & 54.9 & 65.0 & 68.8 & 94.5 & 15.96 & 35.8 \\
    & GT + Gaussian (std=20.0) & 1 & 59.5 & 48.5 & 58.5 & 54.4 & 63.4 & 67.3 & 94.3 & 31.92 & 7.5 \\
    \hline
    \multirow{2}{*}{\makecell[c]{Ablation (iii)}} & FlowStep3D (epoch=120) & 1 & 36.0 & 24.6 & 35.4 & 26.4 & 53.8 & 53.7 & 57.8 & 12.21 & 72.8 \\
    & \textbf{FlowStep3D (epoch=120)} & \textbf{2} & 54.4 & 42.4 & 52.4 & 47.3 & 58.8 & 63.7 & 93.6 & 2.29 & 96.3 \\
    \hline
  \end{tabular}
  \vspace{-0.4cm}
\end{table}

\begin{figure}[h]
\setlength{\abovecaptionskip}{ 4 pt}
  \centering
  \includegraphics[width=0.95\linewidth]{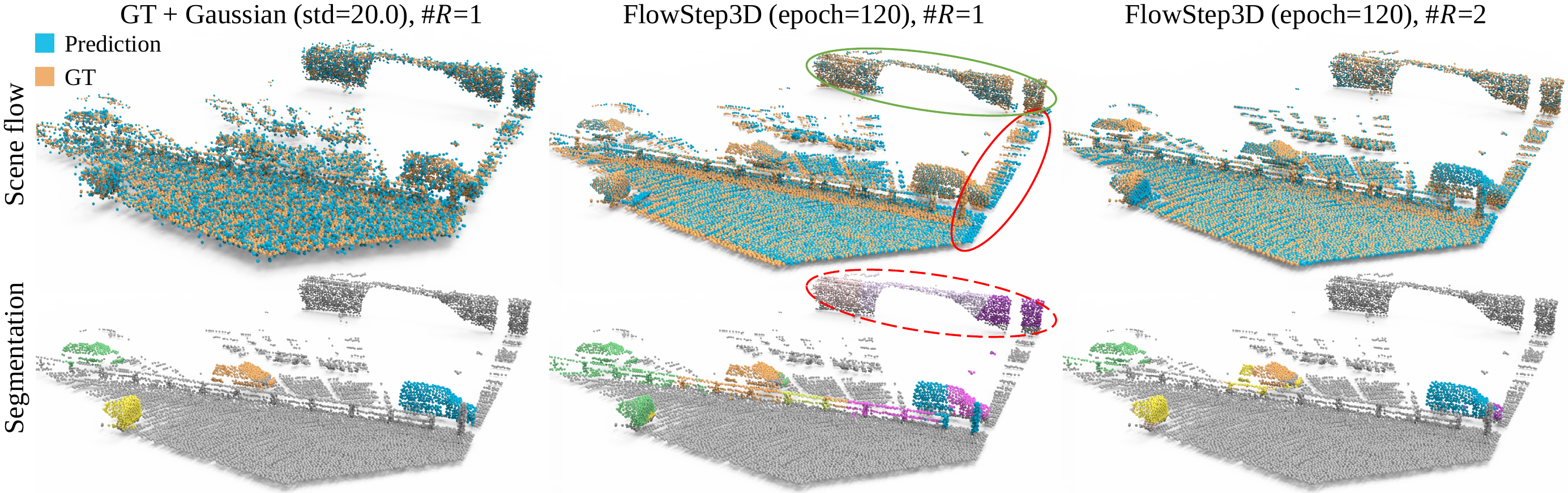}
  \caption{Qualitative results for ablation study about the robustness to scene flow distortions on KITTI-SF. Scene flows are visualized via the point cloud warped by the flows. In the middle column, the scene flow estimations are accurate for the above-ground background points (solid green ellipse) but have biased errors for the ground plane points (which can be clearly seen inside the solid red ellipse). This deviation of scene flow estimations between the above-ground background and the ground plane leads to over-segmentation of the background (dashed red ellipse).}
\label{fig:kittisf_flow}
\vspace{-0.3cm}
\end{figure}

\paragraph{Analysis on KITTI-SF:} As shown in Table \ref{tab:robust_kittisf}, our method has strong robustness to Gaussian noise, same as on OGC-DR dataset. The model achieves 59.5 AP even when the scene flow is corrupted by Gaussian noise to only 7.5 AccR. In contrast, the model without iterative optimization only gives 36.0 AP when the scene flow AccR is 72.8. Figure \ref{fig:kittisf_flow} shows qualitative results. In the middle column, although scene flows have an overall high quality, the inconsistency in scene flows between two parts of the same object leads to over-segmentation. We believe the Weighted Kabsch algorithm inside our dynamic rigid loss is the key. This algorithm inherently smooths the Gaussian-like noise in scene flows but cannot handle the biased errors. Fortunately, our object-aware ICP is designed to correct such inconsistency in the flows, thus improving segmentation performance in iterative optimization.

\phantom{xxx}\textbf{(4) Choice of Hyperparameters for Smoothness Regularization}

\begin{figure}[h]
  \centering
  \includegraphics[width=1.0\linewidth]{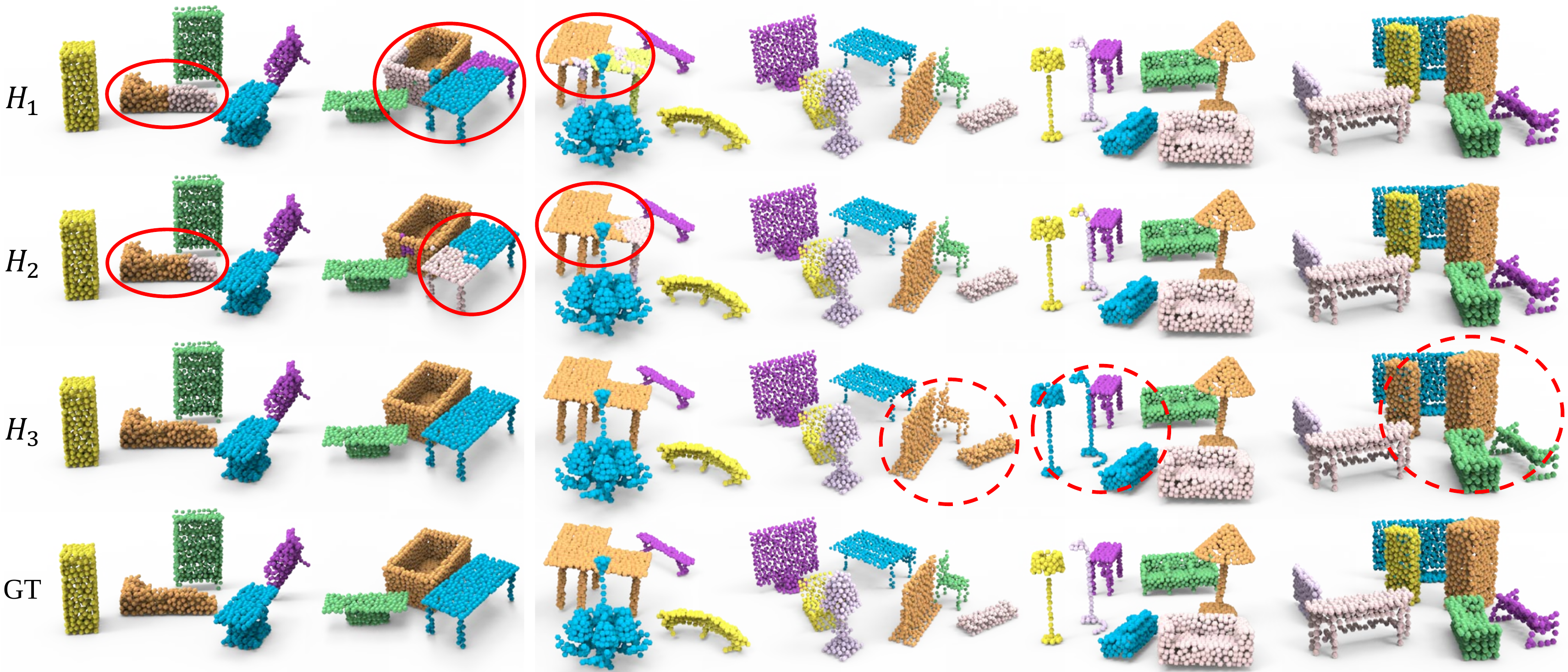}
  \vspace{0.2cm}
  \caption{Qualitative results for the influence of smoothness regularization hyperparamters on OGC-DR. $H_3$ reduces over-segmentation issues in $H_1$/$H_2$ (solid red ellipses in column 1 $\sim$ 3), but fails to separate different objects sometimes (dashed red ellipses in column 4 $\sim$ 6).}
  \vspace{0.1cm}
  \label{fig:ogcdr_smoothH}
 \vspace{-0.4cm}
\end{figure}

We evaluate the influence of smoothness regularization hyperparameters on OGC-DR, as shown in Table \ref{tab:ogcdr_smoothH}. When we strengthen the regularization by enforcing smoothness in a larger local neighborhood (\ie{}, ablation $H_3$), the Precision score improves with less over-segmentation, while the Recall score is sacrificed. Figure \ref{fig:ogcdr_smoothH} shows qualitative results. In general, as expected, such hyperparameters control the trade-off between over- and under-segmentation issues.

\begin{table}[h]
      \tabcolsep= 0.2cm 
  \caption{Different choices of smoothness regularization hyperparamters on the OGC-DR dataset. The hyperparameter $k$ controls the $K$ nearest neighbors selected within a ball with radius $r$.}
  \label{tab:ogcdr_smoothH}
  \centering
  \begin{tabular}{ccccccccc} 
    \toprule
    & $(k_1, r_1), (k_2, r_2)$ & AP$\uparrow$ & PQ$\uparrow$ & F1$\uparrow$ & Pre$\uparrow$ & Rec$\uparrow$ & mIoU$\uparrow$ & RI$\uparrow$ \\ \midrule
    $H_1$ & (4, 0.01), (8, 0.02) & 92.2 & 83.4 & 88.1 & \underline{83.1} & \underline{93.7} & 90.5 & 97.6  \\
    \textbf{$H_2$ (Full OGC)} & (8, 0.02), (16, 0.04) & 92.3 & 85.1 & 89.4 & \underline{85.6} & \underline{93.6} & 90.8 & 97.8 \\
    $H_3$ & (32, 0.08), (64, 0.16) & 84.6 & 81.0 & 87.4 & \underline{89.5} & \underline{85.4} & 82.3 & 96.8 \\ \bottomrule
  \end{tabular}
  \vspace{-0.2cm}
\end{table}

\phantom{xxx}\textbf{(5) Weighted Smoothness Regularization via Motion Similarity}

We investigate a variant of the smoothness regularization which is weighted by the inter-point motion similarity. This motion-similarity-weighted smoothness regularization is mathematically defined as,
\begin{equation}
    \ell_{smooth}' = \frac{1}{N} \sum_{n=1}^N \Big( \frac{1}{H} \sum_{h=1}^H d( \boldsymbol{o}_{p_n}, \boldsymbol{o}_{p_n^h} ) \cdot \frac{exp(- \| \boldsymbol{m}_{p_n} - \boldsymbol{m}_{p_n^h} \|_2 / \tau)}{E} \Big)
\end{equation}
where $\boldsymbol{m}_{p_n} \in \mathbb{R}^{1\times 3}$ represents the motion vector of center point $\boldsymbol{p}_n$, and $\boldsymbol{m}_{p_n^h} \in \mathbb{R}^{1\times 3}$ represents the motion vector of its $h^{th}$ neighbouring point. $\tau=0.01$ is a temperature factor. $E$ is a normalization term, \ie{}, $E = \sum_{h=1}^H exp(- \| \boldsymbol{m}_{p_n} - \boldsymbol{m}_{p_n^h} \|_2 / \tau)$. This variant selectively enforces the object mask smoothness among points with close locations and similar motions. Intuitively, it may avoid blurry predictions on object boundaries.

\begin{table}[h]
      \tabcolsep= 0.2cm 
  \caption{Quantitative results of motion-similarity-weighted smoothness regularization on KITI-SF. Bold text denotes \textbf{the configuration of full OGC.}}
  \label{tab:weight_smooth}
  \centering
  \footnotesize
  \begin{tabular}{cccccccccc} 
    \toprule
    Flow Source & \#$R$ & Regularizer & AP$\uparrow$ & PQ$\uparrow$ & F1$\uparrow$ & Pre$\uparrow$ & Rec$\uparrow$ & mIoU$\uparrow$ & RI$\uparrow$ \\ 
    \midrule
    \textbf{FlowStep3D (epoch=120)} & \textbf{2} & $l_{smooth}$ & 54.4 & 42.4 & 52.4 & 47.3 & 58.8 & 63.7 & 93.6 \\
    FlowStep3D (epoch=120) & 2 & $l_{smooth}'$ & 49.8 & 40.7 & 50.1 & 46.0 & 55.0 & 61.1 & 93.5 \\ 
    \midrule
    GT + Gaussian (std=10.0) & 1 & $l_{smooth}$ & 61.1 & 49.5 & 59.5 & 54.9 & 65.0 & 68.8 & 94.5 \\
    GT + Gaussian (std=10.0) & 1 & $l_{smooth}'$ & 60.0 & 48.1 & 58.5 & 54.0 & 63.9 & 66.9 & 94.5 \\ 
    \midrule
    GT + Gaussian (std=20.0) & 1 & $l_{smooth}$ & 59.5 & 48.5 & 58.5 & 54.4 & 63.4 & 67.3 & 94.3 \\
    GT + Gaussian (std=20.0) & 1 & $l_{smooth}'$ & 57.9 & 46.5 & 56.3 & 51.1 & 62.6 & 67.2 & 94.4 \\
    \bottomrule
  \end{tabular}
  \vspace{-0.2cm}
\end{table}

\textbf{Analysis:} As shown in Table \ref{tab:weight_smooth}, $l_{smooth}'$ brings no benefits to our \nickname{} method under various scene flow situations. We believe the weighting via motion similarity makes $l_{smooth}'$ more sensitive to noises in scene flow estimations, thus being inferior to our $l_{smooth}$.

\subsection{Limitations of \nickname{}}
\label{sec:limit}

Our method can neither segment non-rigid objects nor discover unseen object types due to the lack of supervision signals. 
Besides, the trained \nickname{} model may not provide generalizable intermediate representations to boost the performance of supervised model, which is discussed in details below. 

\begin{table}[tbh]
      \tabcolsep= 0.2cm 
  \caption{Quantitative results of OGC as a pre-training step on KITTI-Det.}
  \label{tab:pretrain_kitti}
  \centering
  \begin{tabular}{rrccccccc}
    \toprule
    training strategy & AP$\uparrow$ & PQ$\uparrow$ & F1$\uparrow$ & Pre$\uparrow$ & Rec$\uparrow$ & mIoU$\uparrow$ & RI$\uparrow$ \\
    \midrule
    train OGC$_{sup}$ on 10\% labelled KITTI-Det & 71.5 & 58.3 & 68.4 & 61.8 & 76.7 & 79.1 & 96.0 \\
    \makecell[c]{(train \nickname{} on unlabelled KITTI-SF \\ + finetune on 10\% labelled KITTI-Det)} & 66.4 & 53.6 & 63.3 & 56.0 & 72.7 & 76.2 & 95.1 \\
    train \nickname{} on unlabeled KITTI-SF & 41.0 & 30.9 & 37.7 & 31.4 & 47.0 & 59.9 & 85.0 \\
    \bottomrule
  \end{tabular}
  \vspace{-0.4cm}
\end{table}

We investigate whether our unsupervised method OGC can be used as a pre-training technique before fully-supervised fine-tuning with a small amount of labeled data, like other popular self-supervised representation learning methods. To do this, we firstly keep a subset (10\%) of labelled point clouds from KITTI-Det training set, and then use our OGC model unsupervisedly trained on KITTI-SF to be fine-tuned on the labelled subset. For comparison, we additionally train a new model with full supervision on the same labelled subset from scratch.

\textbf{Analysis:} From Table \ref{tab:pretrain_kitti}, we can see that: 1) Not surprisingly, using the pre-trained OGC model followed by fine-tuning brings a significant improvement over our unsupervised OGC model, \ie{}, the 1st row $vs$ the 3rd row. 2) However, the (pre-train + fine-tune) strategy fails to outperform the fully-supervised model from scratch, \ie{}, the 1st row $vs$ the 2nd row. 
Fundamentally, this is because our unsupervised OGC is not dedicated to learn general intermediate representations for multiple downstream tasks. Instead, our OGC is task-driven and it aims to directly segment objects from raw point clouds. The learned latent representations in unsupervised training are likely to be different from the latent representations learned in fully-supervised training. In this regard, a na\"ive combination of (pre-train + fine-tune) may confuse the network and give inferior results. Nevertheless, how to effectively leverage the unsupervised model along with full supervision is an interesting direction and we leave it for future exploration.




\subsection{Additional Qualitative Results}
\label{sec:more_qual}

\begin{figure}[t]
\setlength{\abovecaptionskip}{ 4 pt}
  \centering
  \includegraphics[width=1.0\linewidth]{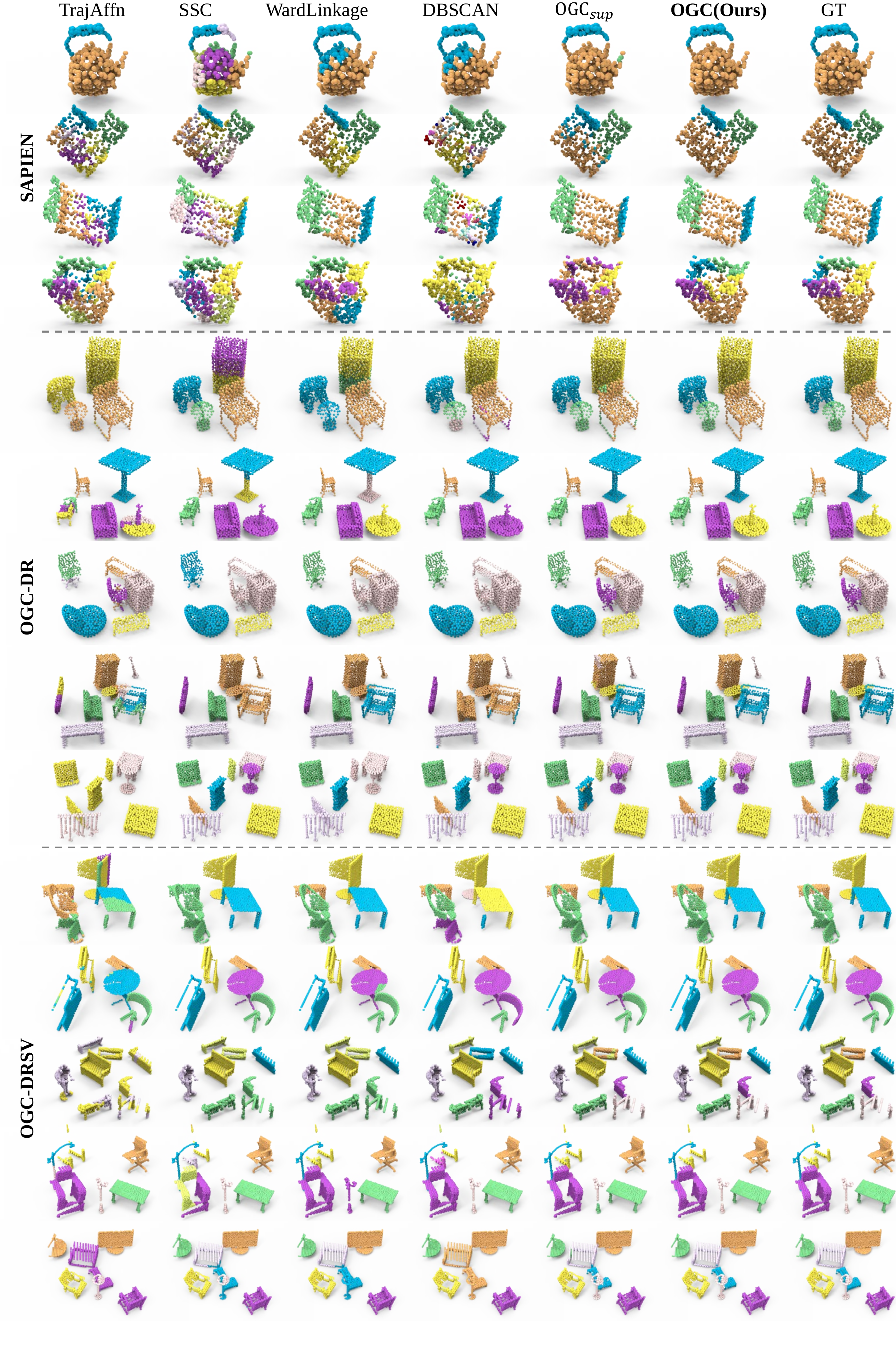}
  \caption{Additional qualitative results on SAPIEN, \nickname{}-DR, and \nickname{}-DRSV.}
\label{fig:qual_appsyn}
\end{figure}

\begin{figure}[t]
\setlength{\abovecaptionskip}{ 4 pt}
  \centering
  \includegraphics[width=1.0\linewidth]{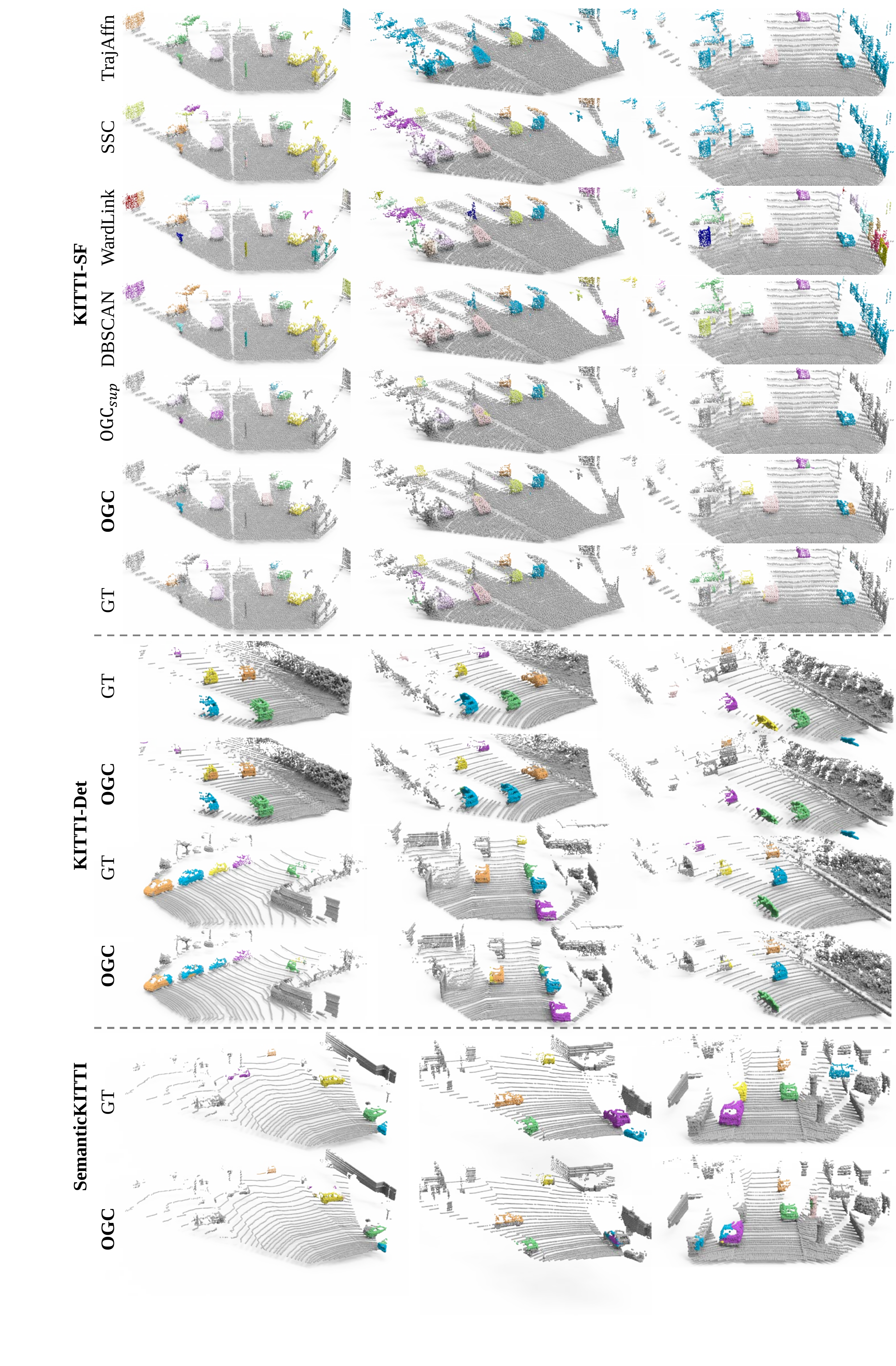}
  \caption{Additional qualitative results on KITTI-SF, KITTI-Det and SemanticKITTI.}
\label{fig:qual_appreal}
\end{figure}

We provide additional qualitative results in Figure \ref{fig:qual_appsyn} for experiments in Sections \ref{sec:eval_sapien} and \ref{sec:eval_indoor} on SAPIEN and \nickname{}-DR datasets, and in Figure \ref{fig:qual_appreal} for experiments in Section \ref{sec:eval_kittisf} and \ref{sec:eval_kittidet} on the KITTI datasets. 

\begin{figure}[h]
\setlength{\abovecaptionskip}{ 4 pt}
  \centering
  \includegraphics[width=1.0\linewidth]{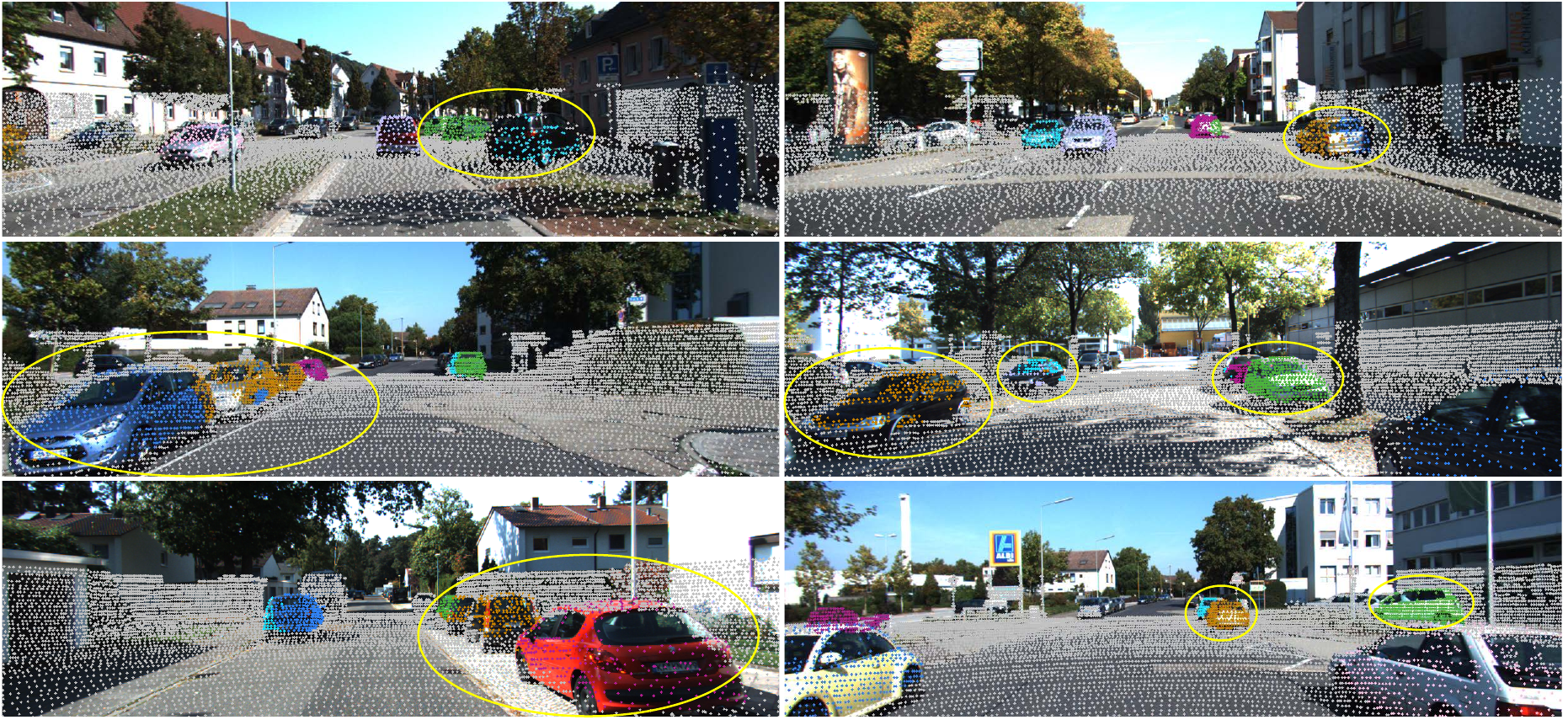}
  \caption{Qualitative results for static object segmentation on KITTI. Images in the 1st row are from KITTI-SF dataset, the rest are from KITTI-Det dataset. Static cars in yellow ellipses are parking alongside the road and can be successfully segmented by our method.}
\label{fig:qual_kitti_static}
\end{figure}

For better visualization, we additionally project the segmented point clouds onto the corresponding RGB images in KITTI-SF and KITTI-Det datasets. As shown in Figure \ref{fig:qual_kitti_static}, our method OGC can successfully segment static cars parking alongside the road, thanks to our geometry invariance loss $l_{invariant}$ which enables our network to generalize the segmentation strategy to similar yet static objects through a set of scene transformations.

